\documentclass[lettersize,journal]{IEEEtran}
\usepackage{amsmath,amsfonts}
\usepackage{algorithmic}
\usepackage{algorithm}
\usepackage{array}
\usepackage[caption=false,font=normalsize,labelfont=sf,textfont=sf]{subfig}
\usepackage{textcomp}
\usepackage{stfloats}
\usepackage{url}
\usepackage{verbatim}
\usepackage{graphicx}
\usepackage{hyperref}
\usepackage{natbib}

\usepackage[utf8]{inputenc} 
\usepackage{booktabs}       
\usepackage{amsfonts}       
\usepackage{nicefrac}       
\usepackage{microtype}      
\usepackage{xcolor}         
\usepackage{adjustbox}
\usepackage{multirow} 
\usepackage{colortbl}
\usepackage{amsmath}
\usepackage{pifont}
\usepackage{caption}
\usepackage{listings}
\usepackage{ragged2e}

\definecolor{dkgreen}{rgb}{0,0.6,0}
\definecolor{gray}{rgb}{0.5,0.5,0.5}
\definecolor{mauve}{rgb}{0.58,0,0.82}
\definecolor{citecolor}{HTML}{0071bc}

\definecolor{Gray}{gray}{0.9}

\begin{document}

\title{Universal Image Restoration Pre-training via \\ Masked Degradation Classification}

\author{JiaKui Hu, Zhengjian Yao, Lujia Jin, Yinghao Chen, Yanye Lu
\thanks{{JiaKui Hu, Zhengjian Yao and Yanye Lu are with Institute of Medical Technology, Peking University Health Science Center, Peking University, Beijing, China, and also with  Biomedical Engineering Department, College of Future Technology, Peking University, Beijing, China, and also with  National Biomedical Imaging Center, Peking University, Beijing, China.}}
\thanks{Lujia Jin is with JIUTIAN Research, Beijing, China.}
\thanks{Yinghao Chen is with the College of Electronic Engineering, National University of Defense Technology, Changsha, China.}
\thanks{Corresponding Authors: Yanye Lu (yanye.lu@pku.edu.cn).}
}

\markboth{Journal of \LaTeX\ Class Files,~Vol.~14, No.~8, August~2021}%
{Shell \MakeLowercase{\textit{et al.}}: A Sample Article Using IEEEtran.cls for IEEE Journals}


\maketitle

\begin{abstract}
This study introduces a Masked Degradation Classification Pre-Training method (MaskDCPT), designed to facilitate the classification of degradation types in input images, leading to comprehensive image restoration pre-training. Unlike conventional pre-training methods, MaskDCPT uses the degradation type of the image as an extremely weak supervision, while simultaneously leveraging the image reconstruction to enhance performance and robustness. MaskDCPT includes an encoder and two decoders: the encoder extracts features from the masked low-quality input image. The classification decoder uses these features to identify the degradation type, whereas the reconstruction decoder aims to reconstruct a corresponding high-quality image. This design allows the pre-training to benefit from both masked image modeling and contrastive learning, resulting in a generalized representation suited for restoration tasks. Benefit from the straightforward yet potent MaskDCPT, the pre-trained encoder can be used to address universal image restoration and achieve outstanding performance. Implementing MaskDCPT significantly improves performance for both convolution neural networks (CNNs) and Transformers, with a minimum increase in PSNR of 3.77 dB in the 5D all-in-one restoration task and a 34.8\% reduction in PIQE compared to baseline in real-world degradation scenarios. It also emergences strong generalization to previously unseen degradation types and levels. In addition, we curate and release the UIR-2.5M dataset, which includes 2.5 million paired restoration samples across 19 degradation types and over 200 degradation levels, incorporating both synthetic and real-world data. The dataset, source code, and models are available at \url{https://github.com/MILab-PKU/MaskDCPT}.
\end{abstract}


\section{Introduction}

\IEEEPARstart{U}{niversal} image restoration is the process of employing a single model to transform low-quality (LQ) images affected by variable, mixed, and real-world degradation into high-quality (HQ) images. In recent work, deep learning-based methods~\citep{airnet,promptir,ai2023multimodal,luo2023controlling,zheng2024selective,guo2024onerestore} have demonstrated superior performance and efficiency in solving universal image restoration compared to traditional techniques~\citep{BM3D,yang2010image}. The prevalent approaches employ degradation representations of LQ images as discriminative prompts for universal image restoration tasks, utilizing elements such as gradients~\citep{ma2020structure}, frequency~\citep{ji2021frequency}, supplementary parameters~\citep{promptir}, and features compressed through large neural networks~\citep{airnet,ai2023multimodal,luo2023controlling,zheng2024selective,wang2024promptrestorer}. These degradation representations subsequently function as prompts for base restoration models, which are either fine-tuned or specifically trained for universal image restoration. Despite achieving high performance through the implementation of precise and effective prompts, these methods do not exploit the latent prior information inherent within the restoration models.

\begin{figure}
\centering
\includegraphics[width=\linewidth]{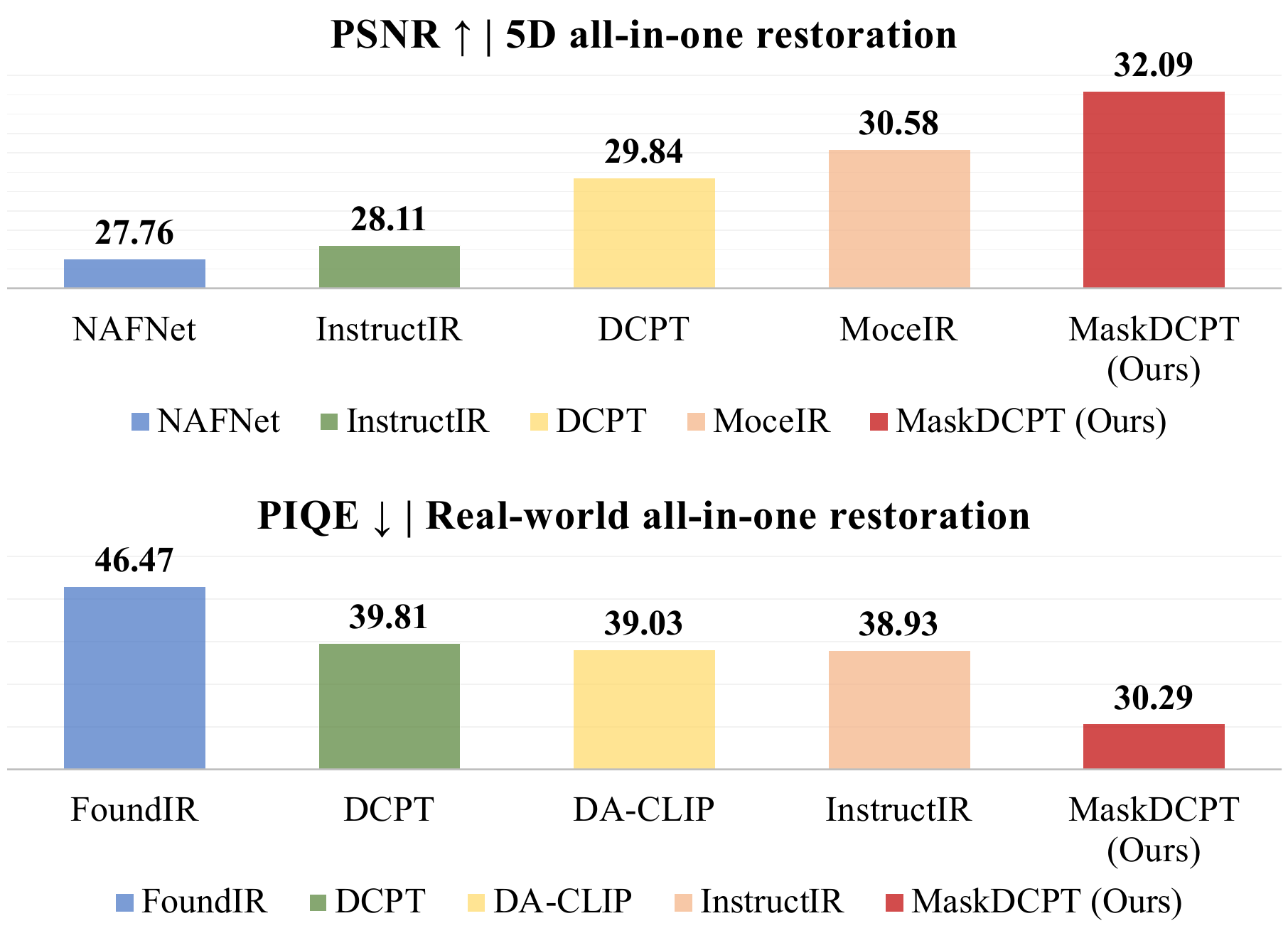}
\caption{MaskDCPT achieves the state-of-the-art fidelity and perception in multiple restoration tasks, including all-in-one and real-world scenarios.}
\label{fig:overall_results}
\vspace{-5mm}
\end{figure}

Pre-training methods~\citep{devlin2018bert,radford2019language,brown2020language,grill2020bootstrap,caron2021emerging,chen2021empirical,he2022masked,xie2022simmim} are adept at exploiting the latent prior information inherent within the restoration models themselves. They can activate latent discriminant information within neural networks, thereby facilitating the acquisition of universal representations and rendering the pre-trained model suitable for downstream tasks. Contrastive learning~\citep{chen2020simple,he2020momentum} discovers representations by maximizing agreement across multiple augmented views of the same sample using contrastive loss~\citep{oord2018representation}, thus obtaining features with fine-grained discriminant information~\citep{chen2021empirical}. Masked Image Modeling (MIM)~\citep{he2022masked,xie2022simmim,tian2022designing} extends BERT's~\citep{devlin2018bert} success from language to vision transformers and CNNs. MIM introduces a challenging image reconstruction task through a substantially high mask ratio, which requires the model to uncover the intrinsic distribution of images. Following GPT's~\citep{radford2019language,brown2020language} success in language generation, related methods~\citep{chen2020generative} are utilized in image generation. PURE~\cite{wei2025perceive} also successfully used pre-trained autoregressive MLLM to adapt to real-world super-resolution. However, pre-training in image restoration~\citep{Liu_2023_DegAE,chen2023masked,qin2024restore} is mainly confined to single-task applications or requires carefully designed fine-tuning methods. This suggests that current approaches do not fully arouse the universal representations provided by extensive pre-training. It is imperative to develop a pre-training framework for restoration models that can handle universal restoration tasks. 



In this paper, we assert that the ability for degradation classification constitutes a frequently overlooked, yet salient, discriminative feature inherent in restoration models. We validate the effectiveness and robustness of neural networks in this capability. First, we examine the degradation classification capabilities of the classical~\citep{liang2021swinir,nafnet,Zamir2021Restormer} and all-in-one~\citep{promptir} image restoration architectures. Models with random initialization possess a preliminary aptitude for degradation classification, which is subsequently refined through all-in-one restoration training, thus enabling a better identification of previously unobserved degradation types. Further investigation reveals that this ability remains intact even when images are randomly masked. This observation indicates that image distribution learning based on masked modeling and degradation distribution learning based on degradation classification can coexist. Drawing upon this finding, we leverage this potential during the pre-training for universal image restoration tasks. By integrating degradation classification, restoration, and reconstruction synergistically during the pre-training phase, the model's proficiency in discerning degradation is significantly enhanced. This methodology not only maintains its efficacy in image restoration, but also fosters a more comprehensive pre-training process.



Building on these insights, we introduce a Masked Degradation Classification Pre-training (\textbf{MaskDCPT}) framework designed for universal image restoration tasks. This approach provides the model with strong prior knowledge about degradation discrimination by simultaneously pre-training three tasks: degradation classification, image reconstruction, and restoration. This enhances the model's ability to identify degradation, supporting the learning of universal restoration representations, and making the pre-trained model suitable for downstream restoration tasks. Specifically, MaskDCPT uses an encoder-decoder structure. The encoder includes an image restoration network without the restoration head. The decoder is divided into two parts: one for degradation classification and the other for image reconstruction and restoration. The encoder transforms the input image into refined latent features. The classification decoder identifies the degradation type of the input image, while the reconstruction decoder, following the MIM design, enables both reconstruction and restoration of the input image using these features. The pre-trained encoder serves as the initialization for the restoration model during fine-tuning, greatly improving restoration performance. Experimental results show that our MaskDCPT framework significantly enhances the effectiveness of various architectures in restoration tasks, including all-in-one, mixed, and real-world degradation scenarios. {Moreover, to accommodate a broad spectrum of degradations present in real-world application scenarios, we curate a dataset consisting of 2.5 million samples, referred to as \textbf{UIR-2.5M}, tailored for the universal image restoration. This dataset covers 19 degradation types and over 200 degradation levels. Experiments indicate that the restoration model trained with the UIR-2.5M dataset demonstrates superior generalization when exposed to unseen degradation.}

In summary, our main contributions are as follows. 

\begin{itemize}
\item We validate that degradation classification is an inherent prior ability of restoration networks. This inherent capability is rapidly enhanced in restoration training and persists even after the input image is masked.
\item We serve the degradation classification as a fundamental component of pre-training. By incorporating it with mask image modeling, we devise the MaskDCPT specifically tailored for universal image restoration.
\item MaskDCPT offers substantial performance gains and can be applied to diverse architectures and tasks. Within the 5D all-in-one restoration task, MaskDCPT achieves a PSNR gain of 4.17, 4.32, 4.38, and 3.77 dB for SwinIR, NAFNet, Restormer, and PromptIR, respectively. When restoring mixed and real-world degradations, MaskDCPT provides a 34.8\% reduction in PIQE to the baseline method.
\item We curate and release the largest universal image restoration dataset, UIR-2.5M. The restoration model trained with UIR-2.5M shows enhanced generalization to unseen degradation types and levels.
\end{itemize}

Compared to our conference paper~\cite{hu2025universal} presented at ICLR 2025, several improvements have been made in this study. The conference version segmented the pre-training into two independent stages, each assigned to degradation classification and generation capacity preservation. In this study, we accomplish the parallelization of them by integrating degradation classification with MIM, thereby enhancing training efficiency and yielding superior pre-training results. This advancement allows us to achieve the State-of-the-art (SoTA) results in 5D all-in-one restoration and mixed degradation tasks. Furthermore, we collect the dataset, UIR-2.5M, for more intricate universal image restoration. Restoration model trained with UIR-2.5M can generalize better in real-world scenarios, unseen degradation types, and levels. Finally, we expand our conference version by incorporating more references and experiments. Compared with recent methods, MaskDCPT consistently delivers superior universal image restoration performance. We further analyze how to improve MaskDCPT's pre-training performance, supported by more ablation studies.

\section{Related Work}

\subsection{Image Restoration}

Recent advances in deep-learning based restoration models~\citep{liang2021swinir,chen2021hinet,Mei_2021_CVPR,Zamir2021Restormer,tu2022maxim,Uformer,nafnet,chen2023hat,jin2024one,hu2025enhancing} have consistently demonstrated superior performance and efficiency compared to traditional techniques in the realm of single task image restoration. The proposed neural networks primarily utilize convolutional neural networks (CNNs)~\citep{lecun2015deep} and Transformers~\citep{vaswani2017attention}. CNNs~\citep{lim2017enhanced,zhang2018rcan,chen2021hinet,Mei_2021_CVPR,nafnet} exhibit exceptional efficacy in processing localized information within images, while Transformers~\citep{liang2021swinir,Zamir2021Restormer,Uformer,chen2023hat,hu2025enhancing} are adept at exploiting the local self-similarity of images through the utilization of long-range dependencies. However, these methods construct specialized models tailored to individual tasks~\citep{nafnet,li2023grl,hu2025enhancing}. Consequently, a significant subset of these techniques proves insufficient to address the inherent diversities associated with image restoration~\citep{li2020all}. 

\textbf{Universal Image Restoration} is conceived for this purpose, which requires a single model to handle various degradations. In early universal restoration approaches, distinct tasks are managed by decoupled learning~\citep{dl} or employing different encoders~\citep{li2020all} or decoder heads~\citep{chen2020pre}. These approaches require the model to explicitly assess degradation types and select distinct network branches to address varied degradations. In recent developments, AirNet~\citep{airnet} employs MoCo~\citep{he2020momentum}, while IDR~\citep{zhang2023ingredient} formulates various physical degradation models to acquire degradation representations for comprehensive image restoration. PromptIR~\citep{promptir} integrates additional parameters via dynamic convolutions to facilitate universal image restoration without recourse to embedded features. DACLIP~\citep{luo2023controlling}, MPerceiver~\citep{ai2023multimodal}, and DiffUIR~\citep{zheng2024selective} harness large external models~\citep{radford2021learning,van2017neural,esser2021taming} or generative priors to achieve improved performance and accommodate more tasks. Furthermore, VLUNet~\cite{zeng2025vision} has advanced the field by developing a deep unfolding network to achieve more stable restoration results. DFPIR~\cite{tian2025degradation} introduces degradation-related parameter perturbations. UniRestore~\cite{chen2025unirestore} introduces considerations for task-oriented image restoration, while UniRes~\cite{zhou2025unires} focuses more on complex mixed degradation. These methods integrate the modulation of external parameters~\citep{promptir}, physical models~\citep{zhang2023ingredient}, human instructions~\citep{conde2024instructir,guo2024onerestore,zeng2025vision}, and the high-dimensional features derived from extensive neural networks~\citep{airnet,ai2023multimodal,luo2023controlling,zheng2024selective,cui2024adair,zhang2025perceive}. However, investigation into the intrinsic potential of the image restoration model and its performance ceiling has been largely overlooked.

\subsection{Pre-training in computer vision}

Pre-training is a way in which intrinsics prior are concealed in input samples and used to improve the performance in downstream tasks. In computer vision, it is divided into two schools: Contrastive Learning (CL)~\citep{chen2020simple,he2020momentum} and Mask Image Modeling (MIM)~\citep{he2022masked,xie2022simmim}. CL aligns features from positive pairs and uniforms the induced distribution of features in the hypersphere~\citep{wang2020understanding}. MIM learns to create before learning to understand~\citep{xie2022simmim}. However, it is difficult to extend to other architectures~\citep{tian2022designing,gao2022mcmae,yao2025scaling} and discards the decoder during downstream tasks, resulting in inconsistent representations between pre-training and fine-tuning~\citep{han2023revcolv}. Recently, many pre-training methods~\citep{chen2020pre,chen2023hat,ijcai2023p121} have been proposed for restoration. Unfortunately, these methods use larger datasets to train larger models in single-degradation settings for pre-training. The existing SSL method~\citep{Liu_2023_DegAE} for image restoration works well in high-cost tasks but is inappropriate for low-cost tasks such as image denoising. RAM~\cite{qin2024restore} pioneered the integration of MIM in the context of all-in-one image restoration. In contrast to them, MaskDCPT's focus lies on examining the influence of masks on the model's capability to discriminate degradation. Using this as a bridge, we aim to effectively merge the learning of degradation classification with the learning of image reconstruction.

\begin{figure*}[!ht]
\centering
\includegraphics[width=\textwidth]{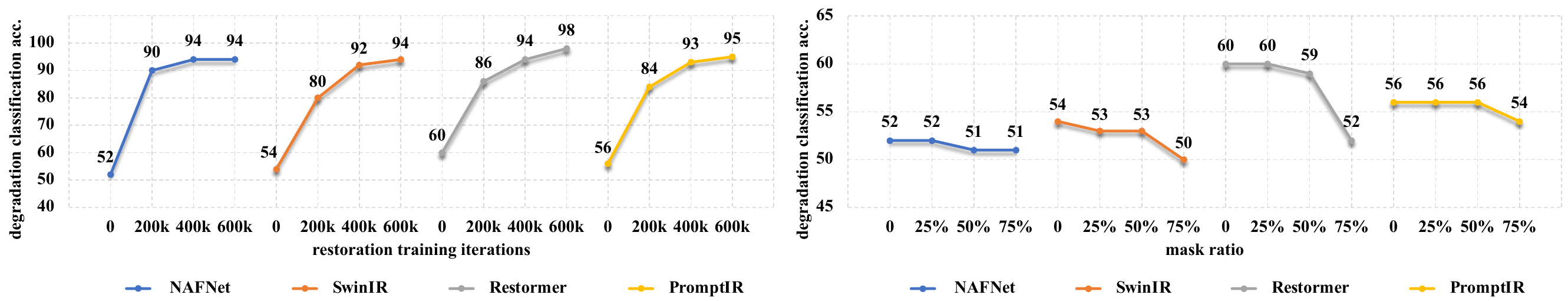}
\caption{Degradation classification accuracy changes with restoration training iterations (left) and image mask ratio (right). The results are averaged under five random seeds.}
\label{fig:motivation}
\vspace{-3mm}
\end{figure*}

\section{Preliminary study}\label{sec:motivation}

Beyond image reconstruction, we discern another significant and foundational ability of image restoration models: degradation classification. We verify that \textit{\textbf{(1)} image restoration models inherently can differentiate between various degradations, \textbf{(2)} there is a degradation classification step in the early training of the restoration model, \textbf{(3)} this ability can be generalized in masked images}. These findings inspire us to believe that optimizing these two intrinsic capabilities may significantly enhance the performance of the model in downstream restoration tasks.

We conduct preliminary experiments to verify these findings, using extracted output features before the restoration head and employing a k-nearest neighbor (kNN) classifier to categorize five degradation types: haze, rain, Gaussian noise, motion blur, and low-light. We randomly select 5,000 images (100 per degradation type) from the datasets: Test1200~\citep{rain100L} for deraining, OTS-BETA~\citep{RESIDE} for dehazing, SIDD~\citep{sidd} for denoising, GoPro~\citep{gopro2017} for deblurring, and LOL~\citep{Chen2018Retinex} for low-light enhancement. The images are center-cropped to maintain uniform feature dimensionality, and then the features are flattened for kNN classification. The dataset is divided into training and test sets in a 2:1 ratio, ensuring an equal distribution of degradation.


\subsection{Inherent degradation classification ability}

As presented in Figure~\ref{fig:motivation} (left) at the next page, random initialized models can achieve the accuracy of the degradation classification of 52 $\sim$ 60 \%. These models inherently possess the capability to classify degradation, highlighting that this is an intrinsic aptitude of neural networks in restoration tasks. 

Drawing inspiration from self-supervised pretraining methods~\cite{he2022masked,chen2020simple,he2020momentum}, it is posited that enhancing this intrinsic capability can lead to improved performance on downstream tasks. The question then arises: \textit{In what ways can this inherent capability of restoration models be optimized?}

\subsection{Degradation classification in restoration training}

We perform an additional verification of the degradation classification capabilities of models trained in three distinct (3D) all-in-one restoration tasks (haze, rain, and Gaussian noise). It should be noted that the five target degradations for classification encompass the three types of degradation used during the training phase. 

The results, as illustrated in Figure~\ref{fig:motivation} (left), indicate that after 3D all-in-one training, the models exhibit an accuracy of 94\% or higher in degradation classification, encompassing degradation types not previously encountered. This result suggests that the all-in-one image restoration training significantly improves the model's ability to classify degradation. Moreover, across four distinct architectures, an increase in the number of restoration training iterations corresponded to an improvement in the model's degradation classification capability. Consequently, during the training of the all-in-one restoration model, while performing restoration tasks, it simultaneously acquires the capability to discern the type of degradation present in the input image.

This experiment demonstrates that \textit{the restoration training can optimize the model's capacity for degradation classification}. It also offers a partial clarification on the success of IPT~\cite{chen2021pre}. The direct employment of the all-in-one restoration task for pre-training serves to enhance the model's capability in degradation classification, thereby substantially augmenting the performance on downstream tasks.

\subsection{Degradation classification in masked images}

We further investigate the robustness of the degradation classification capability of the restoration network when subjected to corrupted input images. Given its prevalence and simplicity of implementation, random masking is employed to simulate image damage. The results are illustrated in Figure~\ref{fig:motivation} (right). When the mask ratio is kept below 50\%, the ability to classify degradation remains largely unchanged. However, a higher mask ratio leads to a reduced classification capability. 

This finding indicates that the degradation classification capability of the restoration model remains notably robust, even in scenarios where the input image is masked. Inspired by MIM~\cite{he2022masked}, image reconstruction can be integrated through the application of a high-ratio mask combined with degradation classification during pre-training. This approach facilitates the learning of the image distribution and concurrently increases the degradation classification for subsequent restoration tasks, thereby enhancing the model's restoration performance.


\section{Method}

\begin{figure*}[!ht]
\centering
\includegraphics[width=0.95\linewidth]{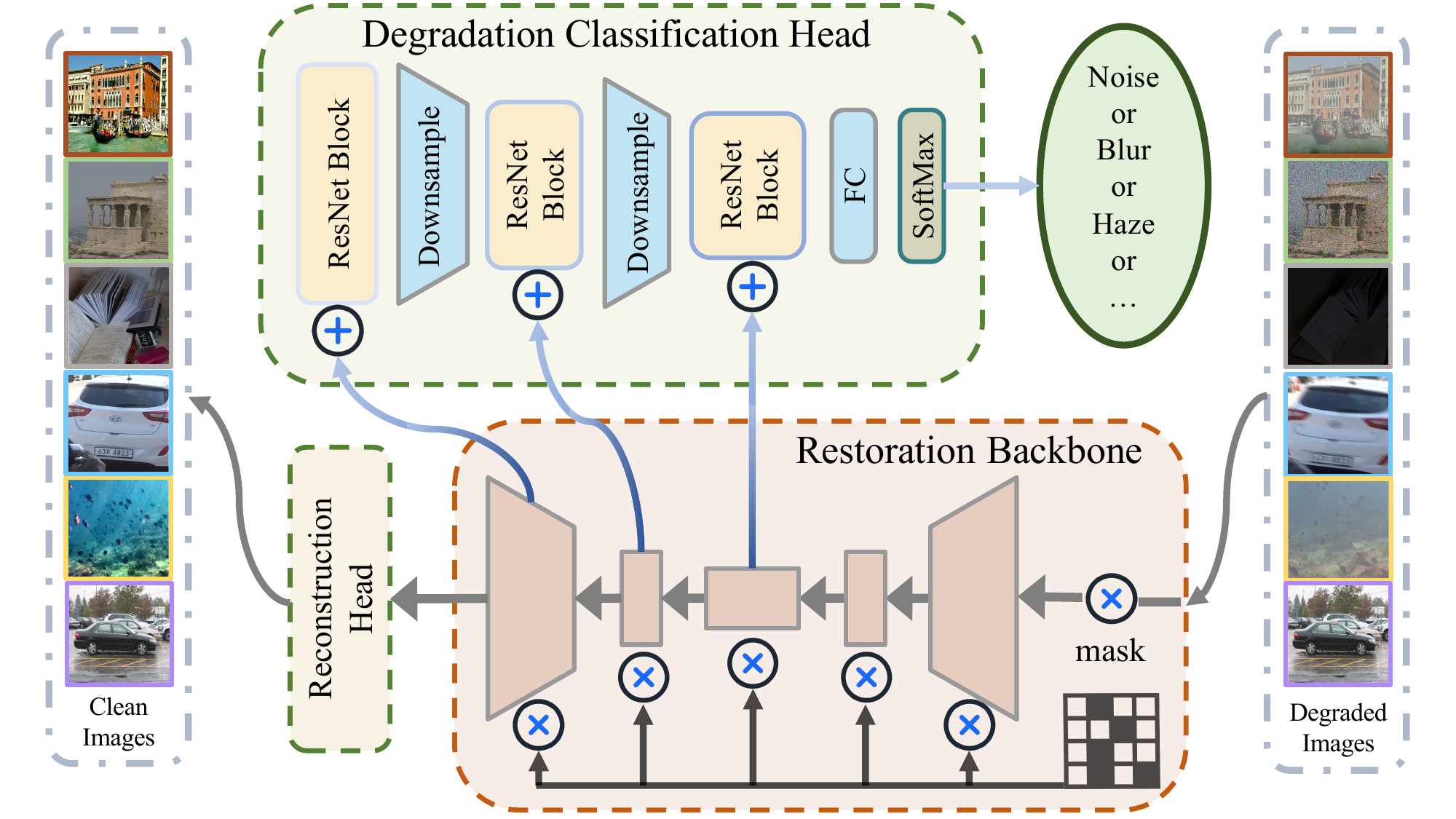}
\caption{MaskDCPT's overall pipeline. {First, MaskDCPT receives degraded images and implements a random patch-level masking to them. Subsequently, the restoration backbone processes the masked images. Throughout this phase, network features are masked to impede the local information leakage. These masked features are then directed towards the reconstruction head for the image restoration as well as the degradation classification head for the degradation classification.} After MaskDCPT, the encoder is fine-tuned for downstream restoration tasks.}\label{fig:method}
\vspace{-3mm}
\end{figure*}

Based on aforementioned analysis, we propose the Masked Degradation Classification Pre-Training (MaskDCPT). We first introduce its overall pipeline and then introduce its specific components. Finally, we introduce the UIR-2.5M dataset that we collected, which includes 19 degradation types commonly seen in real life and 2.5 million images.

\subsection{Overall Pipeline}

MaskDCPT consists of an encoder that comprises restoration models~\citep{liang2021swinir,Zamir2021Restormer,nafnet,promptir} without their restoration heads, and a decoder that classifies the degradation of input images based on the features of the encoder. We leverage Masked Image Modeling (MIM) to facilitate the parallel execution of both the degradation classification and image reconstruction stages. For a given input image $x_{degrad}$ with a specified degradation $D_{gt}$, the image is masking according to a predetermined ratio $r$ and then processed by the encoder to extract the encoder's feature set $F$. Our decoder is equipped with two distinct heads: the reconstruction head, which serves to reconstruct the original image from $F$, and the degradation classification head, which is tasked with determining the type of degradation from $F$. Figure~\ref{fig:method} illustrates this overall pipeline.



\subsection{MaskDCPT}

In this section, we describe the detailed training process of MaskDCPT in components.

\subsubsection{Masked Encoder} Given a low quality image $x_{lq}$, we randomly mask the degraded images (in patch size $16 \times 16$) with a mask ratio $r$. $r$ is 50\% by default.

\begin{equation}
\bar{x}_{lq} = M \odot x_{lq},
\end{equation}

\noindent where $M$ is the mask map and $\odot$ is the Hadamard product.

To achieve a more effective degradation classification, it is crucial to extract features from deeper layers that contain richer high-level semantic information~\citep{cai2022reversible}. However, image restoration models typically adhere to the residual learning design concept~\citep{zhang2017beyond}. The sole reliance on features from the deepest layer for the loss function calculation may result in gradient vanishing in the shallower layers, due to the loss of the encoder's long residual connections~\citep{zhang2017beyond} during feature extraction. To achieve a balance, the features are extracted from each block in the latter part of the encoder. We define these extracted features as $\{F_{i}\}, i \in ({[\frac{l}{2}] + 1, \cdots, l})$, where $l$ is the number of blocks in the network, and $[\cdot]$ is the integer symbol.

\begin{equation}
\{F_{i}\} = \text{Encoder}(\bar{x}_{lq}).
\end{equation}

{When local operators, such as convolution, are used to process input images, these operators integrate surrounding data through the process. This causes the introduction of new signals where the signals should have been masked, leading to information aliasing. To reduce this aliasing~\cite{tian2023designing}, submanifold convolution~\cite{graham2017submanifold} is employed in pretraining.}

\noindent \textbf{Discussion.} {The existing MIM-based restoration pretraining method RAM~\cite{qin2024restore} uses pixel-level mask modeling, which improves the model's ability to capture the distribution of individual pixels. However, this focus on pixel-level details might cause a lack of capturing local information within images. Ablation experiments IX show that, when applying the same fine-tuning strategy, our patch-level mask pretraining outperforms RAM's pixel-level masking approach.}

\subsubsection{Classification Decoder} After extracting masked multi-level features, we feed $\{F_{i}\}$ into the degradation classification decoder (\textbf{\textit{DegCls-Dec}}) of the lightweight decoder to classify the degradation of the input images. The details of the decoder architecture are shown in Figure~\ref{fig:method}. To better aggregate the extracted features, it is necessary to scale up to the features $\{F_{i}\}$. The scaling coefficient $\{\omega_{i}\}$ is learnable. Then, the scaled feature $F'_{i}=\omega_{i}F_{i}$ is plugged into the $i$-th block in ResNet18 to classify the degradation. For stabling the training process, we replace the normalization layers in the decoder from BatchNorm to LayerNorm.

\begin{equation}
\hat{D}_{gt} = \text{DegCls-Dec}(\{F'_i\}).
\end{equation}

It is crucial to note that the challenge of obtaining image restoration data~\citep{li2023lsdir} results in an imbalance in the number of data sets that represent different types of degradation. For example, the deraining dataset Rain200L~\citep{rain200l} comprises only 200 images, whereas the dehazing dataset RESIDE~\citep{RESIDE} encompasses 72,135 images. This imbalance poses a significant long-tail challenge in classifying degradation. To address this issue, we employ Focal Loss~\citep{lin2017focal} as the loss function for long-tail degradation classification.

\begin{equation}
L_{cls} = \text{Focal Loss}(D_{gt}, \hat{D}_{gt}).
\end{equation}

\subsubsection{Reconstruction Decoder} Another MaskDCPT's task is to enable the restoration model to learn clean image distributions by reconstructing masked images. The reconstruction decoder (\textbf{\textit{Recon-Dec}}) in the decoder allows the encoder's feature $F_l$ to reconstruct $\hat{x}_{gt}$, as shown in Figure~\ref{fig:method}. The overall loss function of MaskDCPT is as follows:

\begin{equation}\label{eq:loss}
\begin{aligned}
L_{total} &= \alpha L_{pix} + L_{cls} \\
&= \alpha || x_{gt} - \hat{x}_{gt} ||_{1} + \text{Focal Loss}(D_{gt}, \hat{D}_{gt}),
\end{aligned}
\end{equation}

\noindent where $\alpha$ is 1 by default, and $\hat{x}_{gt} = \text{Recon-Dec}(F_l)$.

\noindent \textbf{Discussion.} Eq.~\ref{eq:loss} performs the simultaneous execution of three tasks: degradation classification, image reconstruction, and image restoration. This is because $L_{pix}$ analyzes both the masked and unmasked regions. The former facilitates image reconstruction from unmasked regions, while the latter aids in the restoration of unblemished images from unmasked areas. As image restoration still enhances the degradation classification, it is posited that the losses in Eq.~\ref{eq:loss} are mutually reinforced, thus fostering an expedited and more robust pre-training.

Furthermore, within masking, $L_{cls}$ in Eq.~\ref{eq:loss} acts as a bridge between MIM and CL. Unlike the ill-posed property of image restoration, degradation classification has a clear and well-defined objective. Under the training objective defined by Eq.~\ref{eq:loss}, the model learns to extract information from partially masked inputs. When images undergo the same degradation but are masked differently, they should still be classified into the same degradation category, indicating a convergence of their learned representations. In contrast, images with different degradations, even if masked by the same mask, should be classified into different degradation categories, reflecting divergent features. This behavior aligns with the core principles of CL.


\subsection{Data collection}

\begin{figure*}[ht]
\centering
\includegraphics[width=\linewidth]{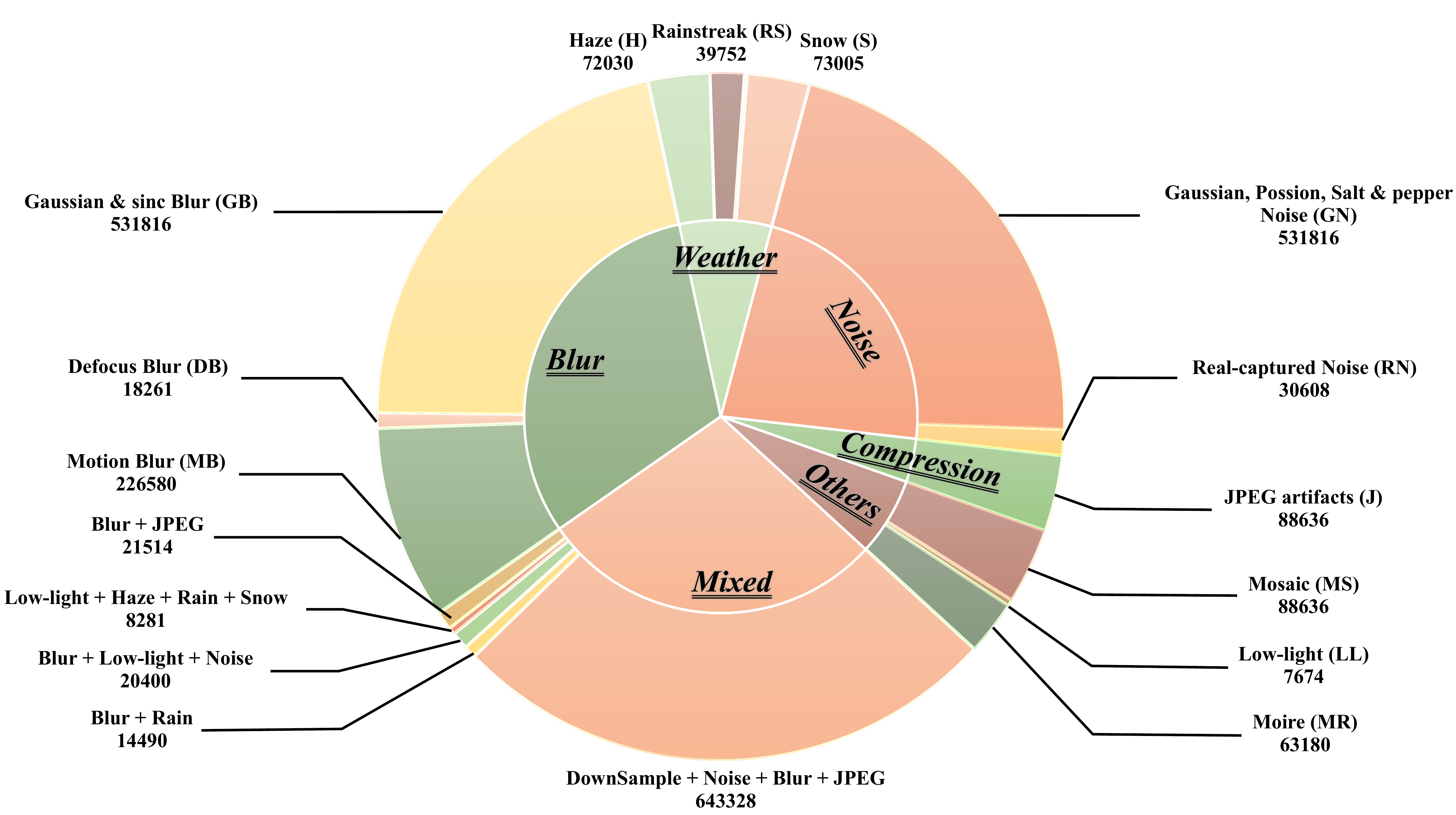}
\caption{{The structure of the UIR-2.5M dataset. It consists of two principal categories, namely single and mixed. Single degradation types contain: blur, weather, noise, compression, and others (suboptimal imaging conditions). Mixed degradation tasks comprise combined distortions resulting from adverse weather, JPEG artifacts, motion blur, and low-light. Our dataset encompasses both synthetic and real-world data instances. The comprehensive dataset includes 2.5 million samples. The detailed distribution of the UIR-2.5M dataset is presented in the Appendix.}}
\label{tab:data}
\vspace{-3mm}
\end{figure*}


To address a comprehensive range of degradations encountered in real-world scenarios, an ideal approach would involve training on a large dataset that includes various degradations and features images rich in texture detail. However, since degraded images and their perfectly registered clean images cannot coexist in real environments, it is a huge challenge to construct a paired general image restoration dataset from real-world data. While generating simulated data is a relatively straightforward task and allows for the creation of complex mixed degradations not easily captured in real-world scenarios, synthetic datasets lack the diversity and realism to effectively train models capable of generalizing to demanding real-world environments. A practical strategy involves curating and filtering existing datasets, followed by preprocessing them into a standardized format conducive to research applications. Therefore, we carefully selected the available training datasets to ensure maximum coverage of different types of degradation and image textures. Table~\ref{tab:data} provides a summary of our curated real and synthetic training datasets, categorized by degradation.

Following the aforementioned operations, a comprehensive collection of 2,482,988 pairs of universal image restoration datasets designated as \textbf{UIR-2.5M} has been assembled, encompassing \textit{single} (1,774,975) and \textit{mixed} (708,013) segments. To enhance applicability in practice, it is noted that in both segments, a proportion of 3\% of the data are sourced from the real-world. Fields such as face, remote sensing, medical imaging, and document remain unexplored and are thus earmarked for future work in the collation of image restoration data within those specific sub-fields. Additionally, local degradation, such as reflection, flare, and incompleteness, has yet to be addressed, with plans to focus on these challenges in future work.


\section{Experiments and Results}\label{sec:experiment}

Our evaluation of MaskDCPT encompasses three distinct scenarios: \textbf{(1) All-in-one.} We fine-tune a single model after MaskDCPT to facilitate image restoration across various degradations, assessing its performance on both 5D all-in-one and 12D tasks. \textbf{(2) Single-task.} Following IDR~\citep{zhang2023ingredient}, we assess the performance of all-in-one trained models in unseen or real-world degradations without fine-tuning. To elucidate the impact of MaskDCPT on single-task pretraining, we present the fine-tuning results of MaskDCPT pretrained models within particular single-task contexts. \textbf{(3) Mixed degradation.} We perform an evaluation of the fine-tuned model under mixed degradation conditions to determine the suitability of MaskDCPT to restore complex degraded images with mixed degradation.

\noindent \textbf{Metrics.} Peak Signal to Noise Ratio (PSNR) and Structural Similarity Index Metric (SSIM) within the sRGB color space are employed to quantify image distortions. In addition, Learned Perceptual Image Patch Similarity (LPIPS) and Fréchet Inception Distance (FID) are utilized as perceptual metrics. For test sets lacking reference HQs, the Perception-based Image Quality Evaluator (PIQE) and the Blind/Referenceless Image Spatial Quality Evaluator (BRISQUE) serve as evaluation metrics. We use the pyiqa~\footnote{https://github.com/chaofengc/IQA-PyTorch} to calculate them.

\begin{table*}[!ht]
\fontsize{6.6pt}{\baselineskip}\selectfont
\setlength\tabcolsep{3pt}
\centering
\begin{tabular}{l|c|c|c|c|c}
\toprule[0.15em]
Method & Dehazing & Deraining & Denoising & Deblurring & Low-Light  \\ 
\midrule[0.15em]
AirNet~\cite{airnet} & 21.04 / 0.884 / 0.077 / 62.52 & 32.98 / 0.951 / 0.058 / 50.12 & 30.91 / 0.882 / 0.102 / 78.12 & 24.35 / 0.781 / 0.189 / 66.13 & 18.18 / 0.735 / 0.122 / 116.9 \\
IDR~\cite{zhang2023ingredient} & 25.24 / 0.943 / 0.052 / 33.25 & 35.63 / 0.965 / 0.043 / 45.62 & 31.60 / 0.887 / 0.092 / 66.24 & 27.87 / 0.846 / 0.178 / 40.83 & 21.34 / 0.826 / 0.108 / 100.6 \\
AdaIR~\cite{cui2024adair} & 30.25 / 0.981 / 0.013 / 13.11 & 37.86 / 0.981 / 0.014 / 13.75 & 31.30 / 0.892 / 0.110 / 46.64 & 28.11 / 0.864 / 0.189 / 19.59 & 22.94 / 0.894 / 0.120 / 52.41 \\  
DA-CLIP~\cite{luo2023controlling} & 29.78 / 0.968 / 0.014 / 15.26 & 35.65 / 0.962 / 0.022 / 22.24 & 30.93 / 0.885 / 0.089 / 54.12 & 27.31 / 0.838 / 0.143 / 23.34 & 21.66 / 0.828 / 0.095 / 55.81 \\ 
RCOT~\cite{tang2024residual} & 30.26 / 0.971 / 0.016 / 16.74 & 36.88 / 0.975 / 0.024 / 19.67 & 31.05 / 0.882 / 0.099 / 62.12 & 28.12 / 0.862 / 0.155 / 21.56 & 22.76 / 0.830 / 0.097 / 61.24 \\  
DA-RCOT~\cite{tang2025degradation} & 30.96 / 0.975 / 0.008 / 10.62 & 37.87 / 0.980 / 0.012 / 12.20 & 31.23 / 0.888 / 0.082 / 37.65 & 28.68 / 0.872 / 0.135 / 12.39 & 23.25 / 0.836 / 0.084 / 47.23 \\ 
MoceIR~\cite{zamfir2025complexity} & 30.72 / 0.979 / 0.013 / 13.28 & 38.01 / 0.982 / 0.014 / 13.63 & 31.34 / 0.893 / 0.103 / 42.93 & 30.04 / 0.901 / 0.143 / 15.11 & 23.00 / 0.902 / 0.118 / 49.77 \\
DFPIR~\cite{tian2025degradation} & 31.23 / 0.982 / 0.013 / 13.48 & 37.56 / 0.979 / 0.016 / 14.71 & 31.26 / 0.892 / 0.091 / 38.75 & 28.79 / 0.879 / 0.164 / 17.07 & 23.79 / 0.895 / 0.122 / 56.35 \\ 
\midrule
SwinIR~\cite{liang2021swinir} & 21.50 / 0.891 / 0.069 / 82.13 & 30.78 / 0.923 / 0.081 / 64.38 & 30.59 / 0.868 / 0.122 / 79.08 & 24.52 / 0.773 / 0.288 / 56.21 & 17.81 / 0.723 / 0.159 / 146.2 \\
\quad + RAM~\cite{qin2024restore} & 28.45 / 0.975 / 0.021 / 10.19 & 26.09 / 0.875 / 0.209 / 92.90 & 31.06 / 0.888 / 0.110 / 39.95 & 26.88 / 0.823 / 0.249 / 36.31 & 21.55 / 0.876 / 0.156 / 89.10 \\ 
\quad + DCPT~\cite{hu2025universal} & {28.68} / {0.977} / 0.019 / 8.93 & {35.70} / {0.975} / 0.022 / 12.10 & \textbf{31.16} / {0.890} / 0.113 / 40.00 & {26.42} / {0.810} / 0.270 / 37.17 & {20.38} / {0.836} / 0.154 / 68.46  \\ 
\quad + \textbf{MaskDCPT (Ours)} & \textbf{29.29} / \textbf{0.981} / \textbf{0.015} / \textbf{5.88} & \textbf{37.16} / \textbf{0.979} / \textbf{0.014} / \textbf{7.60} & 31.13 / \textbf{0.890} / \textbf{0.080} / \textbf{27.41} & \textbf{26.53} / \textbf{0.808} / \textbf{0.218} / \textbf{29.23} & \textbf{21.94} / \textbf{0.905} / \textbf{0.111} / \textbf{55.69} \\ 
\midrule
NAFNet~\cite{nafnet} & 25.23 / 0.939 / 0.053 / 32.68 & 35.56 / 0.967 / 0.050 / 43.57 & 31.02 / 0.883 / 0.139 / 49.57 & 26.53 / 0.808 / 0.206 / 49.12 & 20.49 / 0.809 / 0.141 / 127.9 \\
\quad + {DCPT}~\cite{hu2025universal} & {29.47} / {0.976} / 0.015 / 4.26 & {35.68} / {0.973} / 0.021 / 12.73 & {31.31} / {0.886} / 0.106 / 41.88 & {29.22} / {0.886} / 0.153 / 15.54 & {23.52} / {0.855} / 0.113 / 44.57  \\ 
\quad + \textbf{MaskDCPT (Ours)} & \textbf{31.40} / \textbf{0.978} / \textbf{0.012} / \textbf{3.39} & \textbf{39.92} / \textbf{0.986} / \textbf{0.008} / \textbf{4.21} & \textbf{31.41} / \textbf{0.894} / \textbf{0.076} / \textbf{26.55} & \textbf{31.40} / \textbf{0.920} / \textbf{0.092} / \textbf{7.61} & \textbf{26.31} / \textbf{0.888} / \textbf{0.071} / \textbf{25.88} \\ 
\midrule
Restormer~\cite{Zamir2021Restormer} & 24.09 / 0.927 / 0.067 / 43.62 & 34.81 / 0.962 / 0.050 / 51.69 & \textbf{31.49} / 0.884 / 0.108 / 40.79 & 27.22 / 0.829 / 0.191 / 32.02 & 20.41 / 0.806 / 0.144 / 123.1 \\
\quad + {DCPT~\cite{hu2025universal}} & {29.19} / {0.976} / 0.018 / 6.47 & {36.62} / {0.977} / 0.019 / 11.65 & {31.20} / {0.890} / 0.105 / 44.73 & {28.58} / {0.875} / 0.170 / 18.31 & {23.26} / {0.842} / 0.120 / 59.28  \\ 
\quad + \textbf{MaskDCPT (Ours)} & \textbf{32.67} / \textbf{0.985} / \textbf{0.010} / \textbf{3.12} & \textbf{39.27} / \textbf{0.985} / \textbf{0.009} / \textbf{4.87} & {31.29} / \textbf{0.892} / \textbf{0.076} / \textbf{26.50} & \textbf{30.58} / \textbf{0.910} / \textbf{0.102} / \textbf{11.12} & \textbf{26.11} / \textbf{0.879} / \textbf{0.076} / \textbf{30.33}  \\ 
\midrule
PromptIR~\cite{promptir} & {25.20} / {0.931} / 0.034 / 28.13 & {35.94} / {0.964} / 0.049 / 40.42 & {31.17} / {0.882} / 0.120 / 43.71 & {27.32} / {0.842} / 0.133 / 36.29 & {20.94} / {0.799} / 0.148 / 118.3 \\ 
\quad + {RAM~\cite{qin2024restore}} & 29.63 / 0.975 / 0.014 / 5.64 & 28.11 / 0.888 / 0.178 / 79.38 & 31.08 / 0.889 / 0.109 / 42.79 & 28.00 / 0.862 / 0.183 / 18.36 & 24.45 / 0.907 / 0.120 / 51.45 \\ 
\quad + {DCPT~\cite{hu2025universal}} & {30.93} / {0.982} / 0.012 / 3.89 & {37.18} / {0.979} / 0.016 / 9.75 & {31.27} / {0.891} / 0.110 / 46.10 & {28.86} / {0.880} / 0.164 / 17.61 & {23.09} / {0.840} / 0.128 / 60.42  \\ 
\quad + \textbf{MaskDCPT (Ours)} & \textbf{32.71} / \textbf{0.985} / \textbf{0.009} / \textbf{3.07} & \textbf{39.12} / \textbf{0.985} / \textbf{0.009} / \textbf{4.94} & \textbf{31.30} / \textbf{0.892} / \textbf{0.079} / \textbf{27.67} & \textbf{29.99} / \textbf{0.900} / \textbf{0.111} / \textbf{11.81} & \textbf{26.30} / \textbf{0.881} / \textbf{0.078} / \textbf{29.53}  \\ 
\bottomrule[0.15em]
\end{tabular}
\caption{\small \textbf{\textit{5D all-in-one image restoration results}} in terms of PSNR$\uparrow$ / SSIM$\uparrow$ / LPIPS$\downarrow$ / FID$\downarrow$. Classic restoration models pre-trained with MaskDCPT outperform the methods that require all-in-one specific training and architecture. All methods are trained on widely used 5D all-in-one restoration dataset following IDR~\cite{zhang2023ingredient} to ensure fair comparison.}\label{tab:all_in_one_5d}
\end{table*}

\begin{figure*}[!ht]
\scriptsize
\centering
\resizebox{\textwidth}{!}{
\begin{tabular}{ccc}
\hspace{-0.45cm}
\begin{adjustbox}{valign=t}
\begin{tabular}{c}
\includegraphics[width=0.253\textwidth]{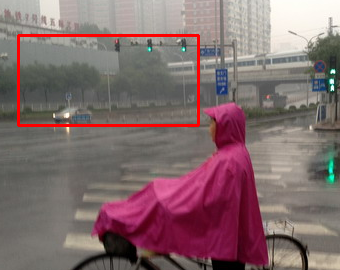}
\vspace{-2pt}
\\
Dehazing (GT)
\end{tabular}
\end{adjustbox}
\hspace{-0.46cm}
\begin{adjustbox}{valign=t}
\begin{tabular}{cccccc}
\includegraphics[width=0.18\textwidth]{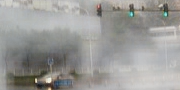} \hspace{-4mm} &
\includegraphics[width=0.18\textwidth]{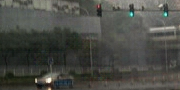} \hspace{-4mm} &
\includegraphics[width=0.18\textwidth]{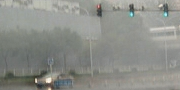} \hspace{-4mm} &
\includegraphics[width=0.18\textwidth]{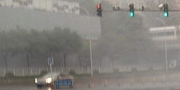} \hspace{-4mm} 
\vspace{-2pt}
\\
SwinIR \hspace{-4mm}  &
NAFNet \hspace{-4mm}  &
Restormer \hspace{-4mm}  &
PromptIR \hspace{-4mm}
\vspace{-1pt}
\\
\includegraphics[width=0.18\textwidth]{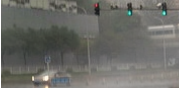} \hspace{-4mm} &
\includegraphics[width=0.18\textwidth]{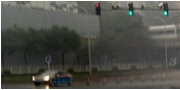} \hspace{-4mm} &
\includegraphics[width=0.18\textwidth]{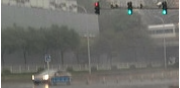} \hspace{-4mm} &
\includegraphics[width=0.18\textwidth]{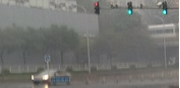} \hspace{-4mm} 
\vspace{-2pt}
\\ 
\textbf{MaskDCPT-SwinIR} \hspace{-4mm} &
\textbf{MaskDCPT-NAFNet} \hspace{-4mm} &
\textbf{MaskDCPT-Restormer}  \hspace{-4mm} &
\textbf{MaskDCPT-PromptIR} \hspace{-4mm}
\vspace{-1pt}
\\
\end{tabular}
\end{adjustbox}
\\
\hspace{-0.45cm}
\begin{adjustbox}{valign=t}
\begin{tabular}{c}
\includegraphics[width=0.253\textwidth]{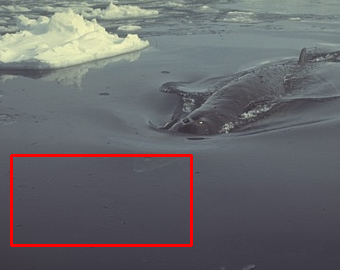}
\vspace{-2pt}
\\
Deraining (GT)
\vspace{-1pt}
\end{tabular}
\end{adjustbox}
\hspace{-0.46cm}
\begin{adjustbox}{valign=t}
\begin{tabular}{cccccc}
\includegraphics[width=0.18\textwidth]{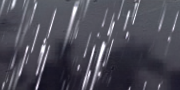} \hspace{-4mm} &
\includegraphics[width=0.18\textwidth]{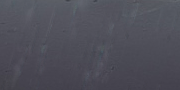} \hspace{-4mm} &
\includegraphics[width=0.18\textwidth]{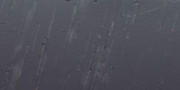} \hspace{-4mm} &
\includegraphics[width=0.18\textwidth]{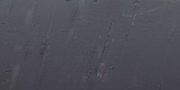} \hspace{-4mm} 
\vspace{-2pt}
\\
SwinIR \hspace{-4mm} &
NAFNet \hspace{-4mm} &
Restormer \hspace{-4mm} &
PromptIR \hspace{-4mm}
\vspace{-1pt}
\\
\includegraphics[width=0.18\textwidth]{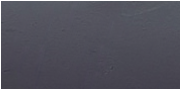} \hspace{-4mm} &
\includegraphics[width=0.18\textwidth]{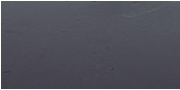} \hspace{-4mm} &
\includegraphics[width=0.18\textwidth]{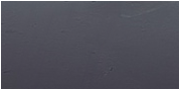} \hspace{-4mm} &
\includegraphics[width=0.18\textwidth]{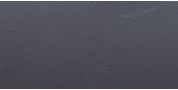} \hspace{-4mm} 
\vspace{-2pt}
\\ 
\textbf{MaskDCPT-SwinIR} \hspace{-4mm} &
\textbf{MaskDCPT-NAFNet} \hspace{-4mm} &
\textbf{MaskDCPT-Restormer}  \hspace{-4mm} &
\textbf{MaskDCPT-PromptIR} \hspace{-4mm}
\vspace{-1pt}
\\
\end{tabular}
\end{adjustbox}
\\
\hspace{-0.45cm}
\begin{adjustbox}{valign=t}
\begin{tabular}{c}
\includegraphics[width=0.253\textwidth]{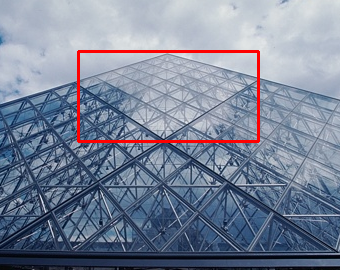}
\vspace{-2pt}
\\
Denoising (GT)
\vspace{-1pt}
\end{tabular}
\end{adjustbox}
\hspace{-0.46cm}
\begin{adjustbox}{valign=t}
\begin{tabular}{cccccc}
\includegraphics[width=0.18\textwidth]{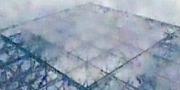} \hspace{-4mm} &
\includegraphics[width=0.18\textwidth]{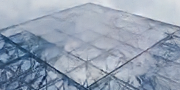} \hspace{-4mm} &
\includegraphics[width=0.18\textwidth]{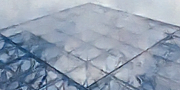} \hspace{-4mm} &
\includegraphics[width=0.18\textwidth]{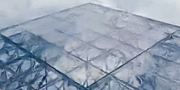} \hspace{-4mm} 
\vspace{-2pt}
\\
SwinIR \hspace{-4mm} &
NAFNet \hspace{-4mm} &
Restormer \hspace{-4mm} &
PromptIR \hspace{-4mm}
\vspace{-1pt}
\\
\includegraphics[width=0.18\textwidth]{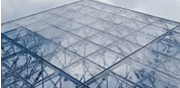} \hspace{-4mm} &
\includegraphics[width=0.18\textwidth]{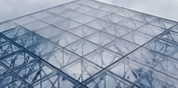} \hspace{-4mm} &
\includegraphics[width=0.18\textwidth]{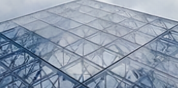} \hspace{-4mm} &
\includegraphics[width=0.18\textwidth]{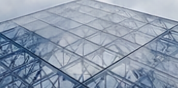} \hspace{-4mm} 
\vspace{-2pt}
\\ 
\textbf{MaskDCPT-SwinIR} \hspace{-4mm} &
\textbf{MaskDCPT-NAFNet} \hspace{-4mm} &
\textbf{MaskDCPT-Restormer}  \hspace{-4mm} &
\textbf{MaskDCPT-PromptIR} \hspace{-4mm}
\vspace{-1pt}
\\
\end{tabular}
\end{adjustbox}
\\
\hspace{-0.45cm}
\begin{adjustbox}{valign=t}
\begin{tabular}{c}
\includegraphics[width=0.253\textwidth]{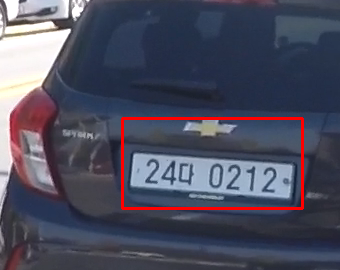}
\vspace{-2pt}
\\
Deblurring (GT)
\vspace{-1pt}
\end{tabular}
\end{adjustbox}
\hspace{-0.46cm}
\begin{adjustbox}{valign=t}
\begin{tabular}{cccccc}
\includegraphics[width=0.18\textwidth]{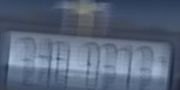} \hspace{-4mm} &
\includegraphics[width=0.18\textwidth]{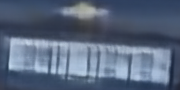} \hspace{-4mm} &
\includegraphics[width=0.18\textwidth]{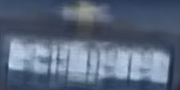} \hspace{-4mm} &
\includegraphics[width=0.18\textwidth]{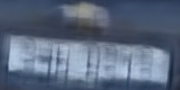} \hspace{-4mm} 
\vspace{-2pt}
\\
SwinIR \hspace{-4mm} &
NAFNet \hspace{-4mm} &
Restormer \hspace{-4mm} &
PromptIR \hspace{-4mm}
\vspace{-1pt}
\\
\includegraphics[width=0.18\textwidth]{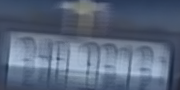} \hspace{-4mm} &
\includegraphics[width=0.18\textwidth]{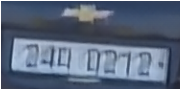} \hspace{-4mm} &
\includegraphics[width=0.18\textwidth]{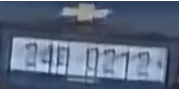} \hspace{-4mm} &
\includegraphics[width=0.18\textwidth]{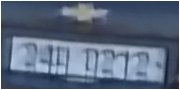} \hspace{-4mm} 
\vspace{-2pt}
\\ 
\textbf{MaskDCPT-SwinIR} \hspace{-4mm} &
\textbf{MaskDCPT-NAFNet} \hspace{-4mm} &
\textbf{MaskDCPT-Restormer}  \hspace{-4mm} &
\textbf{MaskDCPT-PromptIR} \hspace{-4mm}
\vspace{-1pt}
\\
\end{tabular}
\end{adjustbox}
\\
\hspace{-0.45cm}
\begin{adjustbox}{valign=t}
\begin{tabular}{c}
\includegraphics[width=0.253\textwidth]{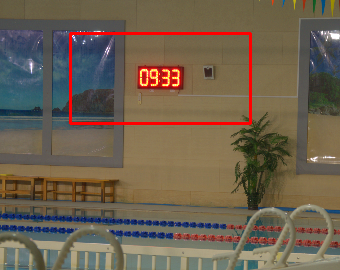}
\vspace{-2pt}
\\
Low-Light (GT)
\vspace{-1pt}
\end{tabular}
\end{adjustbox}
\hspace{-0.46cm}
\begin{adjustbox}{valign=t}
\begin{tabular}{cccccc}
\includegraphics[width=0.18\textwidth]{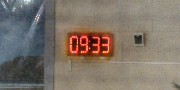} \hspace{-4mm} &
\includegraphics[width=0.18\textwidth]{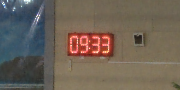} \hspace{-4mm} &
\includegraphics[width=0.18\textwidth]{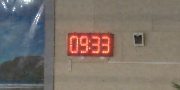} \hspace{-4mm} &
\includegraphics[width=0.18\textwidth]{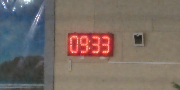} \hspace{-4mm} 
\vspace{-2pt}
\\
SwinIR \hspace{-4mm} &
NAFNet \hspace{-4mm} &
Restormer \hspace{-4mm} &
PromptIR \hspace{-4mm}
\vspace{-1pt}
\\
\includegraphics[width=0.18\textwidth]{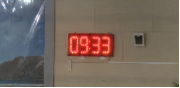} \hspace{-4mm} &
\includegraphics[width=0.18\textwidth]{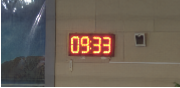} \hspace{-4mm} &
\includegraphics[width=0.18\textwidth]{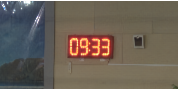} \hspace{-4mm} &
\includegraphics[width=0.18\textwidth]{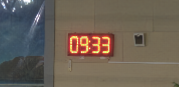} \hspace{-4mm} 
\vspace{-2pt}
\\ 
\textbf{MaskDCPT-SwinIR} \hspace{-4mm} &
\textbf{MaskDCPT-NAFNet} \hspace{-4mm} &
\textbf{MaskDCPT-Restormer}  \hspace{-4mm} &
\textbf{MaskDCPT-PromptIR} \hspace{-4mm}
\vspace{-1pt}
\\
\end{tabular}
\end{adjustbox}
\end{tabular}
}
\caption{\small \textbf{\textit{Visual comparison on 5D all-in-one image restoration datasets.}} Zoom in for best view.}\label{fig:5d_compare}
\vspace{-3mm}
\end{figure*}

\subsection{All-in-one image restoration}

We first assess the performance gain of MaskDCPT on different architectures in all-in-one image restoration.


\noindent \textbf{5D all-in-one dataset.} To facilitate a fair comparison, a subset of UIR-2.5M was meticulously crafted for the 5D all-in-one task. This subset, termed as the \textit{UIR-2.5M-5D} subset, comprises the following: Rain200L, consisting of 200 training images for the purpose of deraining; RESIDE, which includes 72,135 training images alongside 500 test images (SOTS) designated for dehazing; BSD400 and WED, collectively offering 5,144 training images for Gaussian denoising; GoPro, featuring 2,103 training images and 1,111 test images intended for single image motion deblurring; and LOL, which provides 485 training images accompanied by 15 test images for the low-light enhancement.


\noindent \textbf{12D all-in-one dataset.} For the 12D all-in-one task, we use the UIR-2.5M-single for training. In order to comprehensively assess the effectiveness of the restoration model across diverse simulated and real-world conditions, we employ the following datasets for evaluation: MSPFN 5 sets (Rain100L, Rain100H, Test100, Test1200, Test2800), SynRain-13k for deraining; SOTS for dehazing; Snow100K-L for desnowing; RainDS and RainDrop for raindrop removal; LoL v1, LoL v2, LSRW for low-light enhancement; GoPro, HIDE, REDS for deblurring; DPDD for defocus deblurring; Urban100 for Gaussian denoising, deblurring, and demosaicing; SIDD for real-world captured denoising; and RDNet for demoire.

\noindent \textbf{Implementation details.} During MaskDCPT, image restoration models are trained by AdamW optimizer with zero weight decay for 100k iters with batch-size 16 on $256 \times 256$ image patches on 4 NVIDIA L40S GPUs. Due to the heterogeneous encoder-decoder design, we employ different learning rates for the encoder and decoder. The learning rate is set to $3 \times 10^{-4}$ for the encoder and to $1 \times 10^{-4}$ for the decoder. The learning rate does not alter during MaskDCPT. After MaskDCPT, the encoder is used to initialize the image restoration models. 

\noindent \textbf{Dataset sampler.} For degradation with fewer training data, we use repeat sampler technology to ensure that there are enough training pairs for each degradation. For 5D all-in-one image restoration, the repetition ratio is [1H, 300RS, 15GN, 5MB, 60LL]. For 12D all-in-one image restoration, the repetition ratio is [4GN, 4RN, 1MB, 20DB, 1GB, 4J, 5H, 8RS, 180RD, 5S, 4MS, 30LL, 6MR]. The above abbreviations used to represent degradation can be found in the ``abbrev." in Figure~\ref{tab:data}.


\begin{table*}[!ht]
\centering
\fontsize{6.6pt}{\baselineskip}\selectfont
\setlength\tabcolsep{1.15pt}
\begin{tabular}{cccccccccc}
\toprule[0.15em]
\midrule[0.1em] 
\multicolumn{1}{c|}{\multirow{+2}*{Type}} &\multicolumn{1}{c|}{\multirow{+2}*{Method}} & \multicolumn{2}{c|}{\textbf{\textit{Deraining}}} & 
\multicolumn{1}{c|}{\multirow{+2}*{Method}} & \multicolumn{2}{c|}{\textbf{\textit{Dehazing}}} & 
\multicolumn{1}{c|}{\multirow{+2}*{Method}} & \multicolumn{2}{c}{\textbf{\textit{Desnowing}}} \\ 
\cline{3-4} \cline{6-7} \cline{9-10}
\multicolumn{1}{c|}{\multirow{-2}*{Type}} & \multicolumn{1}{c|}{\multirow{-2}*{Method}} &  ~~~PSNR / SSIM $\uparrow$ & \multicolumn{1}{c|}{~~~~~LPIPS / FID $\downarrow$}& 
\multicolumn{1}{c|}{\multirow{-2}*{Method}} &  ~~~PSNR / SSIM $\uparrow$ & \multicolumn{1}{c|}{~~~~~LPIPS / FID $\downarrow$}& 
\multicolumn{1}{c|}{\multirow{-2}*{Method}} & ~~~PSNR / SSIM $\uparrow$ & ~~~~~LPIPS / FID $\downarrow$ \\ 
\midrule[0.1em]
\multicolumn{1}{c|}{\multirow{2}{*}{\begin{tabular}[c]{@{}c@{}}Task \\ Specific\end{tabular}}} & \multicolumn{1}{c|}{DGUNet} 
& 30.99 / 0.913
& \multicolumn{1}{c|}{{0.095 / 32.94}}
& \multicolumn{1}{c|}{Dehamer}
& 34.93 / 0.989
& \multicolumn{1}{c|}{0.008 / 18.11}
& \multicolumn{1}{c|}{SFNet}
& 30.21 / 0.907
& 0.086 / 3.34
\\
\multicolumn{1}{c|}{} & \multicolumn{1}{c|}{Restormer}
& 31.76 / \textbf{0.924}
& \multicolumn{1}{c|}{0.082 / 28.89}
& \multicolumn{1}{c|}{MB-Talor}          
& \textbf{37.54} / \textbf{0.993}
& \multicolumn{1}{c|}{\textbf{0.005} / \textbf{2.545}}
& \multicolumn{1}{c|}{Focal-Net} 
& 29.97 / 0.903
& 0.090 / 3.43
\\
\midrule[0.1em]
\multicolumn{1}{c|}{\multirow{6}{*}{\begin{tabular}[c]{@{}c@{}}All \\ in \\One \end{tabular}}} & \multicolumn{1}{c|}{InsturctIR~\cite{conde2024instructir}} 
& 29.12 / 0.891              
& \multicolumn{1}{c|}{0.100 / 34.98} 
& \multicolumn{1}{c|}{InsturctIR~\cite{conde2024instructir}}  
& 26.90 / 0.952   
& \multicolumn{1}{c|}{0.178 / 44.03}            
& \multicolumn{1}{c|}{InsturctIR~\cite{conde2024instructir}}
& 28.20 / 0.858
& 0.108 / 4.83
\\
\multicolumn{1}{c|}{} & \multicolumn{1}{c|}{DA-CLIP~\cite{luo2023controlling}} 
& 29.95 / 0.902          
& \multicolumn{1}{c|}{0.071 / 24.90} 
& \multicolumn{1}{c|}{DA-CLIP~\cite{luo2023controlling}}  
& 28.19 / 0.965   
& \multicolumn{1}{c|}{0.069 / 8.12}            
& \multicolumn{1}{c|}{DA-CLIP~\cite{luo2023controlling}}
& 28.26 / 0.868
& 0.088 / 2.10
\\
\multicolumn{1}{c|}{} & \multicolumn{1}{c|}{UniRestore~\cite{chen2025unirestore}}
& 27.87 / 0.864         
& \multicolumn{1}{c|}{0.125 / 39.53}          
& \multicolumn{1}{c|}{UniRestore~\cite{chen2025unirestore}} 
& 27.91 / 0.904             
& \multicolumn{1}{c|}{0.093 / 10.35}               
& \multicolumn{1}{c|}{UniRestore~\cite{chen2025unirestore}}
& 28.78 / 0.863
& 0.099 / 3.89
\\
\multicolumn{1}{c|}{} & \multicolumn{1}{c|}{FoundIR~\cite{li2024foundir}}    
& 29.72 / 0.890
& \multicolumn{1}{c|}{0.094 / 33.25}     
& \multicolumn{1}{c|}{FoundIR~\cite{li2024foundir}}        
& 30.12 / 0.974 
& \multicolumn{1}{c|}{0.013 / 4.94}   
&  \multicolumn{1}{c|}{FoundIR~\cite{li2024foundir}}
& 29.64 / 0.895
& 0.082 / 3.36
\\ 
\multicolumn{1}{c|}{} & \multicolumn{1}{c|}{DCPT~\cite{hu2025universal}}
& 30.50 / 0.909
& \multicolumn{1}{c|}{0.077 / 27.25}          
& \multicolumn{1}{c|}{DCPT~\cite{hu2025universal}}           
& 30.40 / 0.976                         
& \multicolumn{1}{c|}{0.015 / 4.29}              
& \multicolumn{1}{c|}{DCPT~\cite{hu2025universal}}
& 30.42 / 0.907
& 0.089 / 3.08
\\
\multicolumn{1}{c|}{} & \multicolumn{1}{c|}{\textbf{MaskDCPT (Ours)}} 
& \textbf{31.93} / {0.922}
& \multicolumn{1}{c|}{\textbf{0.056} / \textbf{20.46}}     
& \multicolumn{1}{c|}{\textbf{MaskDCPT (Ours)}}        
& {31.82} / {0.982} 
& \multicolumn{1}{c|}{{0.011} / {3.19}}   
&  \multicolumn{1}{c|}{\textbf{MaskDCPT (Ours)}}
& \textbf{30.51} / \textbf{0.909}
& \textbf{0.073} / \textbf{1.84}
\\ 
\midrule[0.1em]
\midrule[0.1em] 
\multicolumn{1}{c|}{\multirow{+2}*{Type}} & \multicolumn{1}{c|}{\multirow{+2}*{Method}} & \multicolumn{2}{c|}{\textbf{\textit{Raindrop}}} & 
\multicolumn{1}{c|}{\multirow{+2}*{Method}} & \multicolumn{2}{c|}{\textbf{\textit{Low-light Enhance}}} & 
\multicolumn{1}{c|}{\multirow{+2}*{Method}} & \multicolumn{2}{c}{\textbf{\textit{Motion Deblur}}} \\ 
\cline{3-4} \cline{6-7} \cline{9-10}
\multicolumn{1}{c|}{\multirow{-2}*{Type}}& \multicolumn{1}{c|}{\multirow{-2}*{Method}} &  ~~~PSNR / SSIM $\uparrow$ &\multicolumn{1}{c|}{~~~~~LPIPS / FID $\downarrow$}& 
\multicolumn{1}{c|}{\multirow{-2}*{Method}} &  ~~~PSNR / SSIM $\uparrow$ &\multicolumn{1}{c|}{~~~~~LPIPS / FID $\downarrow$}&  
\multicolumn{1}{c|}{\multirow{-2}*{Method}} &  ~~~PSNR / SSIM $\uparrow$ & ~~~~~LPIPS / FID $\downarrow$               \\ 
\midrule[0.1em] 
\multicolumn{1}{c|}{\multirow{2}{*}{\begin{tabular}[c]{@{}c@{}}Task \\ Specific\end{tabular}}}& \multicolumn{1}{c|}{IDT} 
& 24.55 / 0.803          
& \multicolumn{1}{c|}{0.198 / 56.04} 
& \multicolumn{1}{c|}{GLARE} 
& 19.57 / 0.766         
& \multicolumn{1}{c|}{0.183 / 50.12}
& \multicolumn{1}{c|}{Stripformer} 
& 30.43 / \textbf{0.902}
& \textbf{0.119} / \textbf{8.80}
\\
\multicolumn{1}{c|}{}&\multicolumn{1}{c|}{UDR-S2Former}
& 27.33 / 0.815          
& \multicolumn{1}{c|}{0.340 / 44.38}          
& \multicolumn{1}{c|}{LLFlow-SKF} 
& 22.60 / 0.660      
&  \multicolumn{1}{c|}{0.194 / 58.53}         
& \multicolumn{1}{c|}{DiffIR}
& \textbf{30.53} / 0.898
& 0.128 / 9.76
\\
\midrule[0.1em]
\multicolumn{1}{c|}{\multirow{6}{*}{\begin{tabular}[c]{@{}c@{}}All \\ in \\One \end{tabular}}} & \multicolumn{1}{c|}{InsturctIR~\cite{conde2024instructir}} 
& 21.19 / 0.761           
& \multicolumn{1}{c|}{0.275 / 109.5} 
& \multicolumn{1}{c|}{InsturctIR~\cite{conde2024instructir}}  
& 21.42 / {0.752}   
& \multicolumn{1}{c|}{0.208 / 55.69}            
& \multicolumn{1}{c|}{InsturctIR~\cite{conde2024instructir}}
& 27.85 / 0.847
& 0.194 / 15.43
\\
\multicolumn{1}{c|}{}& \multicolumn{1}{c|}{DA-CLIP~\cite{luo2023controlling}} 
& 24.03 / 0.772
& \multicolumn{1}{c|}{0.180 / 54.91} 
& \multicolumn{1}{c|}{DA-CLIP~\cite{luo2023controlling}}  
& 19.91 / 0.709
& \multicolumn{1}{c|}{0.198 / 52.03}            
& \multicolumn{1}{c|}{DA-CLIP~\cite{luo2023controlling}}
& 26.25 / 0.822
& 0.177 / 15.45
\\
\multicolumn{1}{c|}{} & \multicolumn{1}{c|}{UniRestore~\cite{chen2025unirestore}}
& 20.57 / 0.721
& \multicolumn{1}{c|}{0.333 / 113.2}          
& \multicolumn{1}{c|}{UniRestore~\cite{chen2025unirestore}}         
& 9.55 / 0.276
& \multicolumn{1}{c|}{0.493 / 113.4}               
& \multicolumn{1}{c|}{UniRestore~\cite{chen2025unirestore}}
& 26.29 / 0.807
& 0.194 / 20.95
\\
\multicolumn{1}{c|}{} & \multicolumn{1}{c|}{FoundIR~\cite{li2024foundir}}    
& 21.10 / 0.757
& \multicolumn{1}{c|}{0.275 / 107.3}     
& \multicolumn{1}{c|}{FoundIR~\cite{li2024foundir}}        
& 19.67 / 0.688
& \multicolumn{1}{c|}{0.234 / 50.77}
& \multicolumn{1}{c|}{FoundIR~\cite{li2024foundir}}
& 27.10 / 0.827
& 0.169 / 17.15
\\ 
\multicolumn{1}{c|}{} & \multicolumn{1}{c|}{DCPT~\cite{hu2025universal}}
& 20.32 / 0.751           
& \multicolumn{1}{c|}{0.253 / 108.2}          
& \multicolumn{1}{c|}{DCPT~\cite{hu2025universal}}           
& 19.54 / 0.646                   
& \multicolumn{1}{c|}{0.262 / 59.14}              
& \multicolumn{1}{c|}{DCPT~\cite{hu2025universal}}        
& 27.68 / 0.856
& 0.199 / 16.46
\\
\multicolumn{1}{c|}{}& \multicolumn{1}{c|}{\textbf{MaskDCPT (Ours)}}    
& \textbf{27.57} / \textbf{0.838}
& \multicolumn{1}{c|}{\textbf{0.124} / \textbf{26.83}}     
& \multicolumn{1}{c|}{\textbf{MaskDCPT (Ours)}}        
& \textbf{24.35} / \textbf{0.794}     
& \multicolumn{1}{c|}{\textbf{0.168} / \textbf{34.83}}
& \multicolumn{1}{c|}{\textbf{MaskDCPT (Ours)}}
& 29.83 / 0.884
& 0.127 / 8.68
\\ 
\midrule[0.1em] 
\midrule[0.1em] 
\multicolumn{1}{c|}{\multirow{+2}*{Type}} & \multicolumn{1}{c|}{\multirow{+2}*{Method}} & \multicolumn{2}{c|}{\textbf{\textit{Defocus Deblur}}} & 
\multicolumn{1}{c|}{\multirow{+2}*{Method}} & \multicolumn{2}{c|}{\textbf{\textit{JPEG Removal}}} & 
\multicolumn{1}{c|}{\multirow{+2}*{Method}} & \multicolumn{2}{c}{\textbf{\textit{Real Denoising}}} \\ 
\cline{3-4} \cline{6-7} \cline{9-10}
\multicolumn{1}{c|}{\multirow{-2}*{Type}} & \multicolumn{1}{c|}{\multirow{-2}*{Method}} &  ~~~PSNR / SSIM $\uparrow$
&\multicolumn{1}{c|}{~~~~~LPIPS / FID $\downarrow$}& 
\multicolumn{1}{c|}{\multirow{-2}*{Method}} &  ~~~PSNR / SSIM $\uparrow$
&\multicolumn{1}{c|}{~~~~~LPIPS / FID $\downarrow$}& 
\multicolumn{1}{c|}{\multirow{-2}*{Method}} &  ~~~PSNR / SSIM $\uparrow$ & ~~~~~LPIPS / FID $\downarrow$             \\ 
\midrule[0.1em] 
\multicolumn{1}{c|}{\multirow{2}{*}{\begin{tabular}[c]{@{}c@{}}Task \\ Specific\end{tabular}}}& \multicolumn{1}{c|}{NRKNet} 
& \textbf{26.11} / \textbf{0.817}                  
& \multicolumn{1}{c|}{{0.223 / {43.96}}}
& \multicolumn{1}{c|}{SwinIR} 
& 29.83 / 0.897         
& \multicolumn{1}{c|}{0.084 / 8.20}               
& \multicolumn{1}{c|}{Restormer} 
& \textbf{39.93} / \textbf{0.947}
& 0.198 / 47.24
\\
\multicolumn{1}{c|}{}& \multicolumn{1}{c|}{DRBNet}
& 25.72 / {0.806}          
& \multicolumn{1}{c|}{\textbf{0.182} / {39.37}}          
& \multicolumn{1}{c|}{Restormer} 
& \textbf{32.71} / \textbf{0.960}       
& \multicolumn{1}{c|}{0.043 / 2.90} 
& \multicolumn{1}{c|}{Uformer}               
& 39.80 / {0.946}
& 0.200 / 47.15
\\
\midrule[0.1em] 
\multicolumn{1}{c|}{\multirow{6}{*}{\begin{tabular}[c]{@{}c@{}}All \\ in \\One \end{tabular}}} & \multicolumn{1}{c|}{InsturctIR~\cite{conde2024instructir}} 
& 23.84 / 0.746
& \multicolumn{1}{c|}{0.329 / 84.88} 
& \multicolumn{1}{c|}{InsturctIR~\cite{conde2024instructir}}  
& 31.93 / 0.944   
& \multicolumn{1}{c|}{0.061 / 3.77}            
& \multicolumn{1}{c|}{InsturctIR~\cite{conde2024instructir}}
& 35.45 / 0.881
& 0.356 / 57.45
\\
\multicolumn{1}{c|}{}& \multicolumn{1}{c|}{DA-CLIP~\cite{luo2023controlling}} 
& 23.55 / 0.747 
& \multicolumn{1}{c|}{0.288 / 67.54} 
& \multicolumn{1}{c|}{DA-CLIP~\cite{luo2023controlling}}  
& 30.77 / 0.923
& \multicolumn{1}{c|}{0.079 / 5.58}            
& \multicolumn{1}{c|}{DA-CLIP~\cite{luo2023controlling}}
& 34.18 / 0.838
& {0.301} / {62.47}
\\
\multicolumn{1}{c|}{}&\multicolumn{1}{c|}{UniRestore~\cite{chen2025unirestore}}
& 22.91 / 0.724
& \multicolumn{1}{c|}{0.364 / 91.59}          
& \multicolumn{1}{c|}{UniRestore~\cite{chen2025unirestore}} 
& 30.23 / 0.918
& \multicolumn{1}{c|}{0.080 / 6.40}              
& \multicolumn{1}{c|}{UniRestore~\cite{chen2025unirestore}}
& 35.41 / 0.835
& 0.247 / 56.00
\\
\multicolumn{1}{c|}{}& \multicolumn{1}{c|}{FoundIR~\cite{li2024foundir}}    
& 23.45 / 0.742
& \multicolumn{1}{c|}{0.358 / 89.21}     
& \multicolumn{1}{c|}{FoundIR~\cite{li2024foundir}}       
& 31.43 / 0.930
& \multicolumn{1}{c|}{0.059 / 3.46}   
&  \multicolumn{1}{c|}{FoundIR~\cite{li2024foundir}}
& 37.12 / 0.888
& 0.266 / 46.53
\\ 
\multicolumn{1}{c|}{} & \multicolumn{1}{c|}{DCPT~\cite{hu2025universal}}
& 25.68 / 0.816 
& \multicolumn{1}{c|}{0.216 / 42.59}          
& \multicolumn{1}{c|}{DCPT~\cite{hu2025universal}}           
& 31.89 / 0.947
& \multicolumn{1}{c|}{0.050 / 3.08}             
& \multicolumn{1}{c|}{DCPT~\cite{hu2025universal}}
& 37.07 / 0.881
& 0.282 / 51.03
\\
\multicolumn{1}{c|}{}& \multicolumn{1}{c|}{\textbf{MaskDCPT (Ours)}}    
& 25.64 / 0.809
& \multicolumn{1}{c|}{{0.183} / \textbf{38.49}}     
& \multicolumn{1}{c|}{\textbf{MaskDCPT (Ours)}}       
& 32.02 / 0.944
& \multicolumn{1}{c|}{\textbf{0.039} / \textbf{2.83}}   
&  \multicolumn{1}{c|}{\textbf{MaskDCPT (Ours)}}
& 38.68 / 0.934
& \textbf{0.152} / \textbf{29.48}
\\ 
\midrule[0.1em] 
\midrule[0.1em]
\multicolumn{1}{c|}{\multirow{+2}*{Type}} &\multicolumn{1}{c|}{\multirow{+2}*{Method}} & \multicolumn{2}{c|}{\textbf{\textit{Gaussian Deblur}}} & 
\multicolumn{1}{c|}{\multirow{+2}*{Method}} & \multicolumn{2}{c|}{\textbf{\textit{Demosaic}}} & 
\multicolumn{1}{c|}{\multirow{+2}*{Method}} & \multicolumn{2}{c}{\textbf{\textit{Demoire}}} \\ 
\cline{3-4} \cline{6-7} \cline{9-10}
\multicolumn{1}{c|}{\multirow{-2}*{Type}} & \multicolumn{1}{c|}{\multirow{-2}*{Method}} &  ~~~PSNR / SSIM $\uparrow$ & \multicolumn{1}{c|}{~~~~~LPIPS / FID $\downarrow$}& 
\multicolumn{1}{c|}{\multirow{-2}*{Method}} &  ~~~PSNR / SSIM $\uparrow$ & \multicolumn{1}{c|}{~~~~~LPIPS / FID $\downarrow$}& 
\multicolumn{1}{c|}{\multirow{-2}*{Method}} & ~~~PSNR / SSIM $\uparrow$ & ~~~~~LPIPS / FID $\downarrow$               \\ 
\midrule[0.1em] 
\multicolumn{1}{c|}{\multirow{2}{*}{\begin{tabular}[c]{@{}c@{}}Task \\ Specific\end{tabular}}} & \multicolumn{1}{c|}{SwinIR} 
& 32.91 / 0.918
& \multicolumn{1}{c|}{{0.077 / 2.34}} 
& \multicolumn{1}{c|}{SwinIR} 
& 39.94 / 0.994       
& \multicolumn{1}{c|}{0.006 / 1.03}               
& \multicolumn{1}{c|}{SwinIR}    
& 24.89 / 0.888
& 0.100 / 28.73
\\
\multicolumn{1}{c|}{} & \multicolumn{1}{c|}{Restormer}
& \textbf{33.47} / \textbf{0.930}      
& \multicolumn{1}{c|}{0.064 / 2.21}          
& \multicolumn{1}{c|}{GRL-S}          
& \textbf{41.77} / \textbf{0.996}
& \multicolumn{1}{c|}{\textbf{0.004} / \textbf{0.66}} 
& \multicolumn{1}{c|}{RDNet} 
& \textbf{26.16} / 0.941
& \textbf{0.091} / \textbf{23.64}
\\
\midrule[0.1em]
\multicolumn{1}{c|}{\multirow{6}{*}{\begin{tabular}[c]{@{}c@{}}All \\ in \\One \end{tabular}}}& \multicolumn{1}{c|}{InsturctIR~\cite{conde2024instructir}} 
& 31.37 / 0.884
& \multicolumn{1}{c|}{0.113 / 6.04} 
& \multicolumn{1}{c|}{InsturctIR~\cite{conde2024instructir}}  
& 37.08 / 0.977
& \multicolumn{1}{c|}{0.011 / 2.33}            
& \multicolumn{1}{c|}{InsturctIR~\cite{conde2024instructir}}
& 24.69 / 0.843
& 0.111 / 32.18
\\
\multicolumn{1}{c|}{}& \multicolumn{1}{c|}{DA-CLIP~\cite{luo2023controlling}} 
& 30.89 / 0.867
& \multicolumn{1}{c|}{0.128 / 6.45} 
& \multicolumn{1}{c|}{DA-CLIP~\cite{luo2023controlling}}  
& 38.12 / 0.990
& \multicolumn{1}{c|}{0.006 / 1.07}            
& \multicolumn{1}{c|}{DA-CLIP~\cite{luo2023controlling}}
& 24.75 / 0.826
& 0.134 / 38.71
\\
\multicolumn{1}{c|}{}& \multicolumn{1}{c|}{UniRestore~\cite{chen2025unirestore}}
& 30.77 / 0.871
& \multicolumn{1}{c|}{0.130 / 6.11}          
& \multicolumn{1}{c|}{UniRestore~\cite{chen2025unirestore}} 
& 37.99 / 0.990
& \multicolumn{1}{c|}{0.006 / 1.21}               
& \multicolumn{1}{c|}{UniRestore~\cite{chen2025unirestore}}
& 24.06 / 0.819
& 0.155 / 45.28
\\
\multicolumn{1}{c|}{}& \multicolumn{1}{c|}{FoundIR~\cite{li2024foundir}}    
& 32.90 / 0.915
& \multicolumn{1}{c|}{0.073 / 2.59}     
& \multicolumn{1}{c|}{FoundIR~\cite{li2024foundir}}        
& 38.44 / 0.992
& \multicolumn{1}{c|}{0.005 / 0.84}   
& \multicolumn{1}{c|}{FoundIR~\cite{li2024foundir}}
& 24.71 / 0.876
& 0.107 / 32.49
\\ 
\multicolumn{1}{c|}{} & \multicolumn{1}{c|}{DCPT~\cite{hu2025universal}}
& 32.06 / 0.906 
& \multicolumn{1}{c|}{0.087 / 2.81}          
& \multicolumn{1}{c|}{DCPT~\cite{hu2025universal}}           
& 38.11 / 0.991 
& \multicolumn{1}{c|}{0.006 / 1.05}              
& \multicolumn{1}{c|}{DCPT~\cite{hu2025universal}}
& 24.18 / 0.815
& 0.159 / 31.38
\\
\multicolumn{1}{c|}{}& \multicolumn{1}{c|}{\textbf{MaskDCPT (Ours)}}    
& 33.28 / 0.927
& \multicolumn{1}{c|}{\textbf{0.057} / \textbf{2.13}}     
& \multicolumn{1}{c|}{\textbf{MaskDCPT (Ours)}}        
& 38.59 / 0.992
& \multicolumn{1}{c|}{0.005 / 0.72}   
&  \multicolumn{1}{c|}{\textbf{MaskDCPT (Ours)}}
& 25.21 / \textbf{0.942}
& 0.095 / 24.41
\\ 
\midrule[0.1em]
\bottomrule[0.15em]
\end{tabular}
\caption{\small \textbf{\textit{12D all-in-one image restoration results}} in terms of PSNR$\uparrow$ / SSIM$\uparrow$ / LPIPS$\downarrow$ / FID$\downarrow$. All-in-one network pre-trained with MaskDCPT outperforms task-specific methods in terms of fidelity for deraining, desnowing, raindrop removal, and low-light enhancement. In most restoration tasks, it surpasses task-specific methods in terms of perceptual metrics. All the all-in-one methods are trained on UIR-2.5M to ensure fair comparison.}\label{tab:12d}
\vspace{-3mm}
\end{table*}

\noindent \textbf{5D all-in-one image restoration results} are reported in Table~\ref{tab:all_in_one_5d} and Figure~\ref{fig:5d_compare}. \textbf{(1)} The models pre-trained using MaskDCPT exhibit superior performance compared to the specifically designed all-in-one image restoration architecture across most tasks. In the low-light enhancement task, MaskDCPT-NAFNet achieves an improvement of 2.52 dB over DFPIR and 3.37 dB over AdaIR in terms of PSNR. Additionally, it outperforms IDR~\cite{zhang2023ingredient} by 3.53 dB and DA-RCOT~\cite{tang2025degradation} by 2.72 dB in the motion deblurring task. \textbf{(2)} Regarding the multi-stage training method (IDR), MaskDCPT consistently demonstrates performance improvements. When Restormer~\cite{Zamir2021Restormer} serves as the baseline, the performance gain achieved by IDR is confined to 0.74 dB relative to its base method, whereas MaskDCPT provides an average performance gain of 4.38 dB. \textbf{(3)} MaskDCPT exhibits adaptability to a wide range of architectures. Observations indicate that regardless of whether the network employs a CNN or Transformer architecture, and whether it follows a linear~\cite{liang2021swinir} or UNet-like structure~\cite{nafnet,Zamir2021Restormer,promptir}, MaskDCPT consistently delivers an average performance enhancement of 3.77 dB and above in the 5D all-in-one image restoration task. \textbf{(4)} MaskDCPT demonstrates superiority over existing pre-training methods. Compared to PromptIR~\cite{promptir} models pre-trained using the MaskDCPT and RAM~\cite{qin2024restore} frameworks, those pre-trained with MaskDCPT show significant performance improvements. Specifically, in the dehazing task, MaskDCPT-PromptIR surpasses RAM-PromptIR by 3.08 dB and DCPT-PromptIR by 1.78 dB, respectively.

\noindent \textbf{12D all-in-one image restoration results.} Furthermore, the degradation types are scaled up to 12 to determine the efficacy of MaskDCPT in the presence of a greater number of degradation types. Following DACLIP~\citep{luo2023controlling} and InstructIR~\citep{conde2024instructir}, we use NAFNet as the basic restoration model due to its precision. The performance of the restoration model under 12 degradation is presented in Table~\ref{tab:12d}. It can be observed that, \textbf{(1)} compared to abstract CLIP embeddings~\citep{luo2023controlling}, complex human instructions~\citep{conde2024instructir}, and the large diffusion model~\citep{li2024foundir,chen2025unirestore}, the degradation classification prior to the MaskDCPT-trained model is more effective in addressing the complex all-in-one restoration task. In the context of motion deblurring, the MaskDCPT framework demonstrates an improvement of 1.98 dB, 3.58 dB, and 3.54 dB in PSNR metrics compared to InsturctIR, DA-CLIP, and UniRestore, respectively, while achieving a reduction of 50\% in FID metrics. Moreover, MaskDCPT achieves state-of-the-art performance across all other assessed all-in-one tasks. \textbf{(2)} The restoration model trained with MaskDCPT demonstrates superior performance over previous task-specific methods in terms of both fidelity and perceptual quality. For instance, in desnowing task, MaskDCPT surpasses FocalNet by 0.54 dB in PSNR; in low-light enhancement, it exceeds GLARE by 4.78 dB in PSNR. MaskDCPT also exhibits advances over task-specific approaches in perceptual assessments. For the real image denoising task, MaskDCPT achieves a 37.4\% reduction in the FID compared to Uformer; and in the raindrop removal task, it obtains a 63.5\% reduction in LPIPS relative to UDR-S2Former. \textbf{(3)} The universal restoration method performs similarly to task-specific approaches under global degradation. However, for non-uniform degradation such as haze, motion blur, or defocus blur, task-specific methods perform better.

\begin{table*}[ht]
\centering
\fontsize{6.6pt}{\baselineskip}\selectfont
\setlength\tabcolsep{2.5pt}
\begin{tabular}{l|ccc|cc|ccc|cc|ccc|cc}
\toprule[0.15em]
\multirow{3}{*}{Method} & \multicolumn{5}{c|}{Urban100} & \multicolumn{5}{c|}{Kodak24}  & \multicolumn{5}{c}{BSD68} \\ 
\specialrule{0em}{1pt}{1pt}
\cline{2-16}
\specialrule{0em}{1pt}{1pt}
~ & \multicolumn{3}{c|}{ID} & \multicolumn{2}{c|}{OOD} & \multicolumn{3}{c|}{ID} & \multicolumn{2}{c|}{OOD} & \multicolumn{3}{c|}{ID} & \multicolumn{2}{c}{OOD} \\ 
& $\sigma=15$ & $\sigma=25$ & $\sigma=50$ & $\sigma=60$ & $\sigma=75$ 
& $\sigma=15$ & $\sigma=25$ & $\sigma=50$ & $\sigma=60$ & $\sigma=75$ 
& $\sigma=15$ & $\sigma=25$ & $\sigma=50$ & $\sigma=60$ & $\sigma=75$\\ 
\midrule[0.15em]
AdaIR~\cite{cui2024adair} & 34.10 & 31.68 & 28.28 & 26.63 & 22.60 & 34.88 & 32.39 & 29.22 & 27.39 & 23.15 & 34.01 & 31.34 & 28.06 & 26.47 & 22.83 \\ 
DA-RCOT~\cite{tang2025degradation} & 33.95 & 31.29 & 26.36 & 22.03 & 16.83 & 34.73 & 31.96 & 26.93 & 21.82 & 16.23 & 33.84 & 30.91 & 25.95 & 21.62 & 16.29 \\ 
MoceIR~\cite{zamfir2025complexity} & 33.99 & 31.58 & 28.21 & 26.86 & 23.45 & 34.85 & 32.37 & 29.20 & 27.72 & 23.84 & 33.98 & 31.34 & 28.06 & 26.65 & 23.26 \\
DFPIR~\cite{tian2025degradation} & 33.94 & 31.59 & 28.29 & 26.08 & 21.45 & 34.77 & 32.32 & 29.20 & 26.66 & 21.38 & 33.94 & 31.29 & 28.05 & 25.85 & 21.28 \\ 
\midrule
SwinIR-5D & 32.79 & 30.18 & 26.52 & 24.47 & 19.80 & 33.89 & 31.32 & 27.93 & 25.36 & 20.01 & 33.31 & 30.59 & 27.13 & 24.39 & 20.11 \\
\quad + {RAM~\cite{qin2024restore}} & 33.77 & \textbf{31.41} & \textbf{27.95} & 24.93 & 20.56 & 34.51 & 32.10 & 28.90 & 26.03 & 20.85 & 33.63 & 31.06 & 27.80 & 24.95 & 20.82 \\ 
\quad + {DCPT~\cite{hu2025universal}} & {33.64} & {31.14} & {27.63} & 24.36 & 20.04 & {34.63} & {32.11} & {28.86} & 25.98 & 20.54 & \textbf{33.82} & \textbf{31.16} & \textbf{27.86} & 24.49 & 20.29 \\ 
\quad + \textbf{MaskDCPT (Ours)} & \textbf{33.83} & 31.39 & 27.91 & \textbf{25.63} & \textbf{21.31} & \textbf{34.65} & \textbf{32.15} & \textbf{28.93} & \textbf{26.15} & \textbf{21.39} & 33.78 & 31.13 & 27.85 & \textbf{25.51} & \textbf{21.25} \\ 
\midrule
NAFNet-5D & 33.14 & 30.64 & 27.20 & 25.74 & 19.93 & 34.27 & 31.80 & 28.62 & 25.92 & 18.08 & 33.67 & 31.02 & 27.73 & 25.90 & 19.42 \\
\quad + {DCPT~\cite{hu2025universal}} & {33.64} & {31.23} & {27.98} & 26.30 & 20.13 & {34.72} & {32.28} & {29.21} & 27.09 & 19.99 & {33.94} & {31.31} & {28.12} & 26.32 & 19.84 \\ 
\quad + \textbf{MaskDCPT (Ours)} & \textbf{34.11} & \textbf{31.80} & \textbf{28.63} & 27.03 & 20.79 & \textbf{34.92} & \textbf{32.49} & \textbf{29.42} & 27.49 & 20.31 & \textbf{34.03} & \textbf{31.41} & \textbf{28.21} & 26.58 & 20.17 \\ 
\quad + \textbf{MaskDCPT-12D (Ours)} & 33.86 & 31.49 & 28.23 & \textbf{27.28} & \textbf{25.92} & 34.75 & 32.31 & 29.17 & \textbf{28.28} & \textbf{26.89} & 33.91 & 31.29 & 28.04 & \textbf{27.16} & \textbf{25.86} \\ 
\midrule
Restormer-5D & 33.72 & 31.26 & 28.03 & 25.98 & 21.89 & 34.78 & 32.37 & 29.08 & 26.91 & 23.68 & 34.03 & 31.49 & 28.11 & 25.31 & 22.97 \\
\quad + {DCPT~\cite{hu2025universal}} & {34.14} & {31.79} & \textbf{28.58} & 26.31 & 22.12 & \textbf{34.96} & \textbf{32.49} & \textbf{29.40} & 27.34 & 24.08 & \textbf{34.09} & \textbf{31.46} & \textbf{28.25} & 26.33 & 23.77 \\ 
\quad + \textbf{MaskDCPT (Ours)} & \textbf{34.17} & \textbf{31.81} & 28.53 & \textbf{26.94} & \textbf{23.81} & 34.83 & 32.36 & 29.20 & \textbf{27.57} & \textbf{24.30} & 33.91 & 31.29 & 28.06 & \textbf{26.60} & \textbf{23.93}  \\ 
\midrule
PromptIR-5D & 33.27 & 30.85 & 27.41 & 25.74 & 19.22 & 34.44 & 31.95 & 28.71 & 26.53 & 19.41 & 33.85 & 31.17 & 27.89 & 24.49 & 19.14 \\
\quad + {{RAM~\cite{qin2024restore}}} & 33.64 & 31.27 & 27.90 & 25.94 & 19.65 & 34.46 & 32.01 & 28.81 & 26.96 & 19.80 & 33.68 & 31.08 & 27.81 & 24.96 & 19.84 \\ 
\quad + {{DCPT~\cite{hu2025universal}}} & {33.88} & {31.49} & {28.15} & 26.71 & 22.90 & {34.78} & {32.30} & {29.14} & \textbf{27.58} & 23.52 & \textbf{33.96} & \textbf{31.32} & {28.08} & 25.93 & 21.77 \\ 
\quad + {\textbf{MaskDCPT (Ours)}} & \textbf{34.14} & \textbf{31.79} & \textbf{28.52} & \textbf{26.91} & \textbf{23.84} & \textbf{34.81} & \textbf{32.34} & \textbf{29.20} & 27.56 & \textbf{24.51} & 33.92 & 31.30 & \textbf{28.08} & \textbf{26.62} & \textbf{23.91} \\ 
\bottomrule[0.15em]
\end{tabular}
\caption{\small \textbf{[ZS] \textit{Gaussian denoising results}} in five levels, including in-domain ($\sigma=(15, 25, 50)$) and out-of-domain ($\sigma=(60, 75)$) degradation levels. MaskDCPT improves performance for in-domain (ID) degradation levels. With scaling degradation types and levels in training data, the restoration model can generalize better to out-of-domain (OOD) degradation levels.}
\label{tab:gaussion_denoising}
\end{table*}

\begin{figure*}[ht]
\centering
\setlength{\tabcolsep}{2pt}
\begin{tabular}{cccccc}
\includegraphics[width=0.16\textwidth]{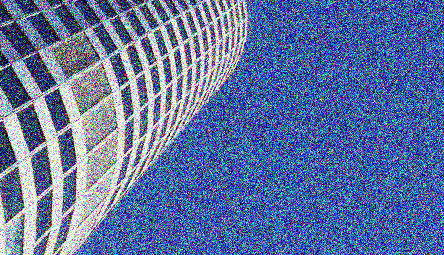} & 
\includegraphics[width=0.16\textwidth]{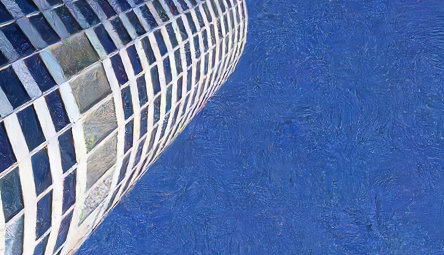} & 
\includegraphics[width=0.16\textwidth]{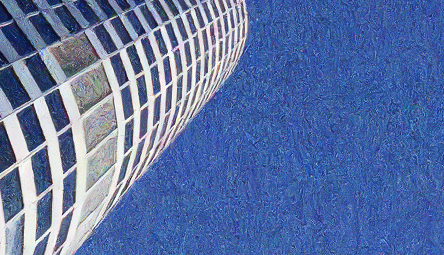} & 
\includegraphics[width=0.16\textwidth]{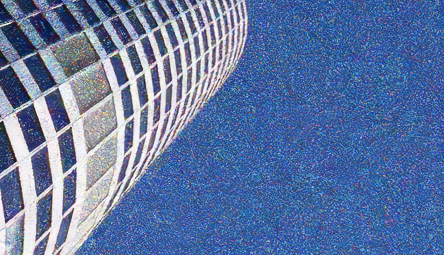} & 
\includegraphics[width=0.16\textwidth]{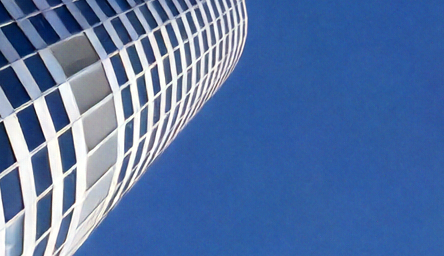} & \includegraphics[width=0.16\textwidth]{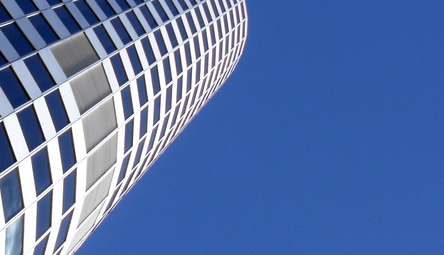} 
\\
Input (LQ) & MoceIR~\cite{zamfir2025complexity} & DFPIR~\cite{tian2025degradation} & \textbf{MaskDCPT-5D} & \textbf{MaskDCPT-12D} & Target (HQ) \\
&&&\textbf{(Ours)}&\textbf{(Ours)}
\end{tabular}
\caption{\small \textbf{\textit{Visual comparison on out-of-domain (OOD) scenarios (Gaussian denoising, $\sigma=75$).}} The MaskDCPT-12D is the only method that effectively removes noise while avoiding the introduction of extraneous artifacts.}\label{fig:ood_dn}
\vspace{-3mm}
\end{figure*}

\subsection{Single-task image restoration}

A further analysis is conducted to determine the suitability of MaskDCPT for single-task image restoration pre-training from two perspectives. \textbf{\romannumeral1. Zero-shot (ZS)}: This evaluates whether MaskDCPT trained models under \textit{12D all-in-one} fine-tuning are used to solve single tasks without optimization. \textbf{\romannumeral2. Fine-tuning (FT)}: This assesses whether the model weights pre-trained with MaskDCPT can be directly used for fine-tuning on single-task image restoration. 

\noindent \textbf{[ZS] Implementation details.} In zero-shot (\textbf{ZS}) settings, we evaluate the performance of the all-in-one models pre-trained with MaskDCPT on the following: {\textbf{(1)} trained tasks at unseen degradation levels}, specifically Gaussian denoising in Urban100, Kodak24, and BSD68 datasets. {\textbf{(2)} Unseen degradation type within unseen real-world scenarios}, including RealBlur-R for motion-debluring, CUHK and PixelDP for defocus-debluring, RealRain1K for deraining, Snow100k-real for desnowing, RTTS for dehazing, along with DICM, LIME, MEF, NPE, and VV for low-light enhancement. These real-world datasets have no reference HQ data.

\noindent \textbf{[FT] Implementation details.} In fine-tuning (\textbf{FT}) configurations, we train the Restormer~\citep{Zamir2021Restormer} model using the Rain13K dataset for image deraining and the GoPro dataset for single image motion deblurring, facilitating a fair comparison with DegAE~\citep{Liu_2023_DegAE}. The training hyperparameters utilized remain consistent with those employed by Restormer~\citep{Zamir2021Restormer}. The key variation lies in the utilization of MaskDCPT pre-trained parameters for model initialization. The fine-tuning process is executed on a single NVIDIA A100 GPU.

\noindent \textbf{[ZS] Unseen degradation levels: Gaussian denoising}. Table~\ref{tab:gaussion_denoising} and Figure~\ref{fig:ood_dn} elucidates the Gaussian denoising results of the image restoration model pre-trained with MaskDCPT across various noise levels, including those degradations not encountered during the training phase. \textbf{(1)} The model pre-trained with MaskDCPT evidences substantial improvements across all architectures and testsets, with a particular emphasis on the high-resolution dataset Urban100~\citep{Urban100}. Specifically, MaskDCPT-SwinIR exhibits an enhancement of 1.39 dB over SwinIR in Gaussian denoising with $\sigma=50$. \textbf{(2)} MaskDCPT displays a marked superiority over existing pre-training methods. Compared to the PromptIR models pre-trained by MaskDCPT and RAM, those pre-trained with MaskDCPT exhibit significant performance enhancements. Notably, within the $\sigma=50$ and the high-resolution dataset Urban100~\citep{Urban100}, MaskDCPT-PromptIR exceeds RAM-PromptIR by 0.97 dB and DCPT-PromptIR by 0.2 dB, respectively. \textbf{(3)} Following exposure to a broader spectrum of degradations, the model demonstrates considerable progress in addressing unseen synthesized levels. In particular, MaskDCPT-NAFNet-12D outperforms MaskDCPT-NAFNet-5D by 5.69 dB in unseen Gaussian noise coefficients, \textit{e.g.}, 75. This performance is attributed to the diverse noise types included within UIR-2.5M, which augment the model's ability to comprehend and mitigate complex noise phenomena.

\begin{table*}[ht]
\centering
\fontsize{6.6pt}{\baselineskip}\selectfont
\setlength\tabcolsep{2pt}
\begin{tabular}{l|c|c|c|c|c|c}
\toprule[0.15em]
Degradation & Motion Blur & Defocus Blur & Rain & Snow & Haze & Low-light \\
\midrule[0.1em]
Method & PSNR$\uparrow$ / SSIM$\uparrow$ / LPIPS$\downarrow$ / FID$\downarrow$ & PIQE$\downarrow$ / BRISQUE$\downarrow$ & PSNR$\uparrow$ / SSIM$\uparrow$ / LPIPS$\downarrow$ / FID$\downarrow$ & PIQE$\downarrow$ / BRISQUE$\downarrow$ & PIQE$\downarrow$ / BRISQUE$\downarrow$ & PIQE$\downarrow$ / BRISQUE$\downarrow$ \\
\midrule[0.15em]
DA-CLIP~\cite{luo2023controlling} & 12.54 / 0.280 / 0.471 / 89.81 & 39.48 / 34.98 & 33.24 / 0.939 / 0.102 / 63.87 & 31.34 / 24.45 & 47.67 / 34.90 & 37.64 / 27.45 \\
InstructIR~\cite{conde2024instructir} & \textbf{34.07} / \textbf{0.948} / \textbf{0.079} / 24.04 & 35.31 / 32.21 & 35.80 / 0.964 / 0.096 / 63.83 & 33.35 / 24.41 & 50.97 / 31.45 & 36.08 / 26.31 \\
UniRestore~\cite{chen2025unirestore} & 30.89 / 0.878 / 0.125 / 42.07 & 31.15 / \textbf{27.53} & 32.54 / 0.905 / 0.141 / 91.49 & 32.69 / 27.16 & 46.88 / \textbf{30.95} & 34.63 / 27.05 \\
FoundIR~\cite{li2024foundir} & 29.12 / 0.832 / 0.146 / 43.46 & 47.40 / 41.74 & 36.02 / 0.967 / 0.093 / 64.36 & 33.18 / 26.20 & 61.14 / 42.26 & 44.17 / 33.51 \\
DCPT-NAFNet~\cite{hu2025universal} & 24.78 / 0.772 / 0.195 / 53.92 & 38.75 / 37.71 & 32.82 / 0.914 / 0.159 / 79.27 & 32.59 / 25.02 & 52.40 / 37.97 & 35.48 / 26.97 \\
\textbf{MaskDCPT-NAFNet (Ours)} & 32.21 / 0.907 / 0.090 / \textbf{22.41} & \textbf{28.61} / 30.19 & \textbf{37.02} / \textbf{0.978} / \textbf{0.070} / \textbf{57.37} & \textbf{30.06} / \textbf{23.27} & \textbf{33.98} / 33.21 & \textbf{28.50} / \textbf{24.76} \\
\bottomrule[0.15em]
\end{tabular}
\caption{\small \textbf{[ZS] \textit{Real-world restoration results}} in six real-world degradation types.}
\label{tab:real_world}
\end{table*}

\begin{figure*}[ht]
\centering
\setlength{\tabcolsep}{2pt}
\begin{tabular}{ccccc}
\includegraphics[width=0.18\textwidth]{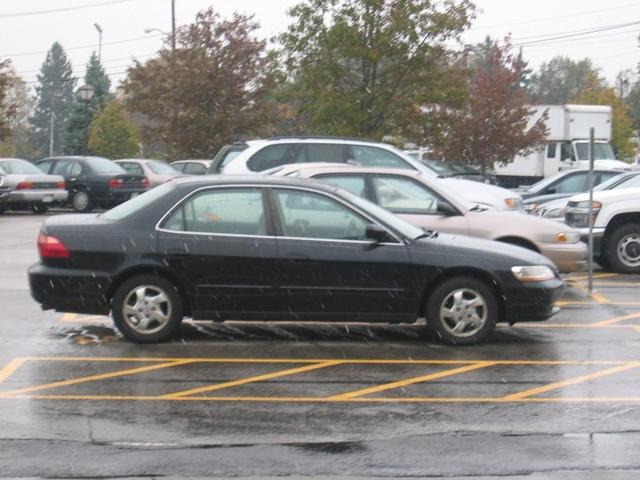} & 
\includegraphics[width=0.18\textwidth]{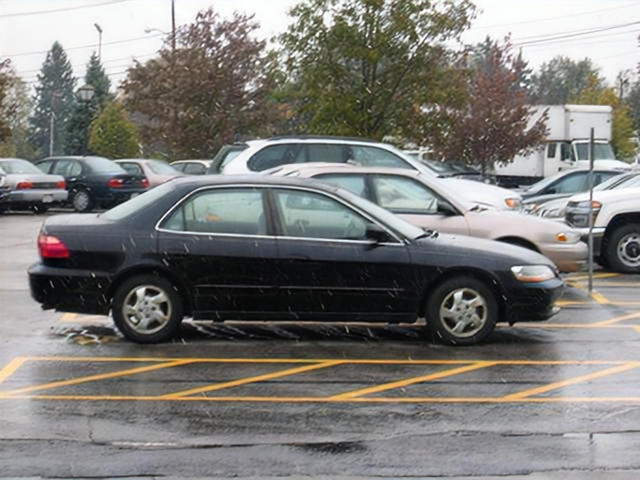} &
\includegraphics[width=0.18\textwidth]{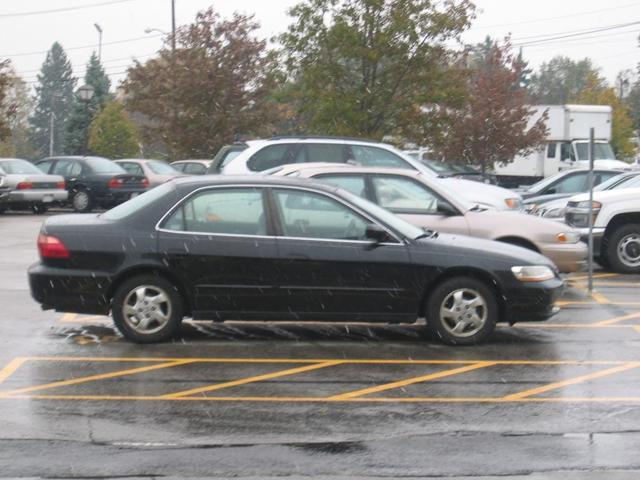} &
\includegraphics[width=0.18\textwidth]{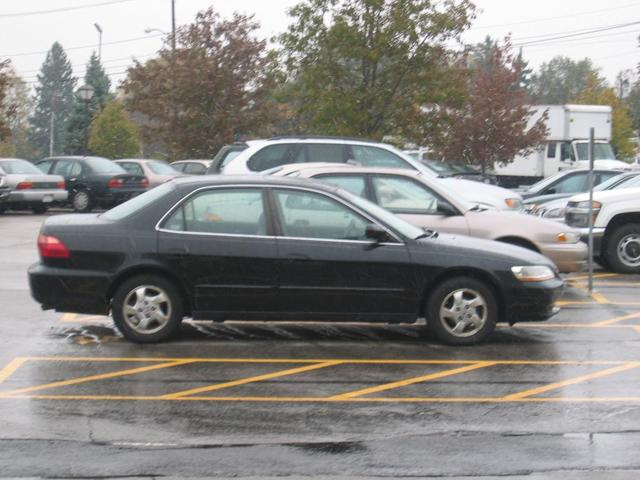} &
\includegraphics[width=0.18\textwidth]{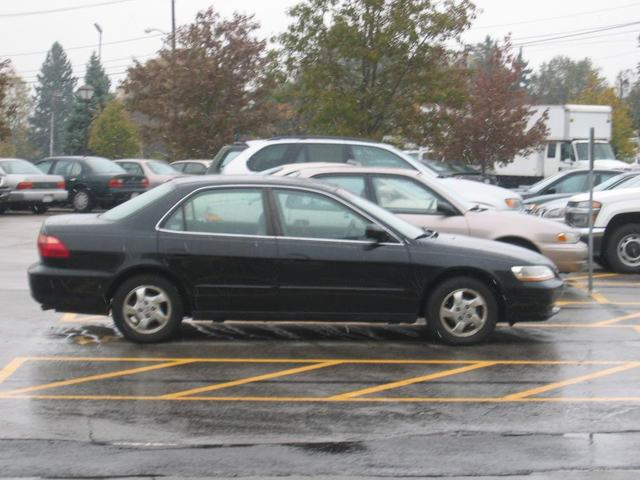} 
\\
\includegraphics[width=0.18\textwidth]{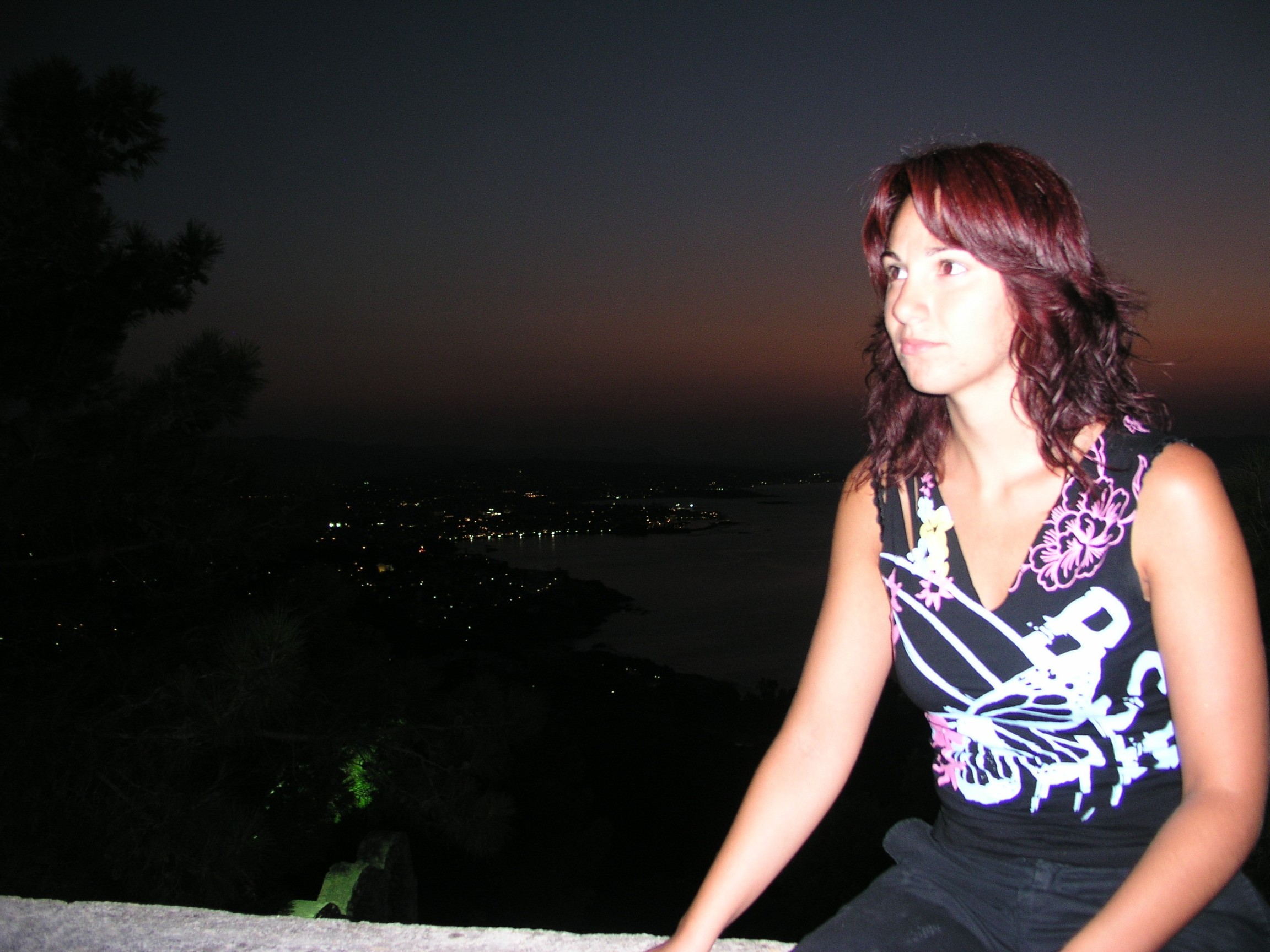} & 
\includegraphics[width=0.18\textwidth]{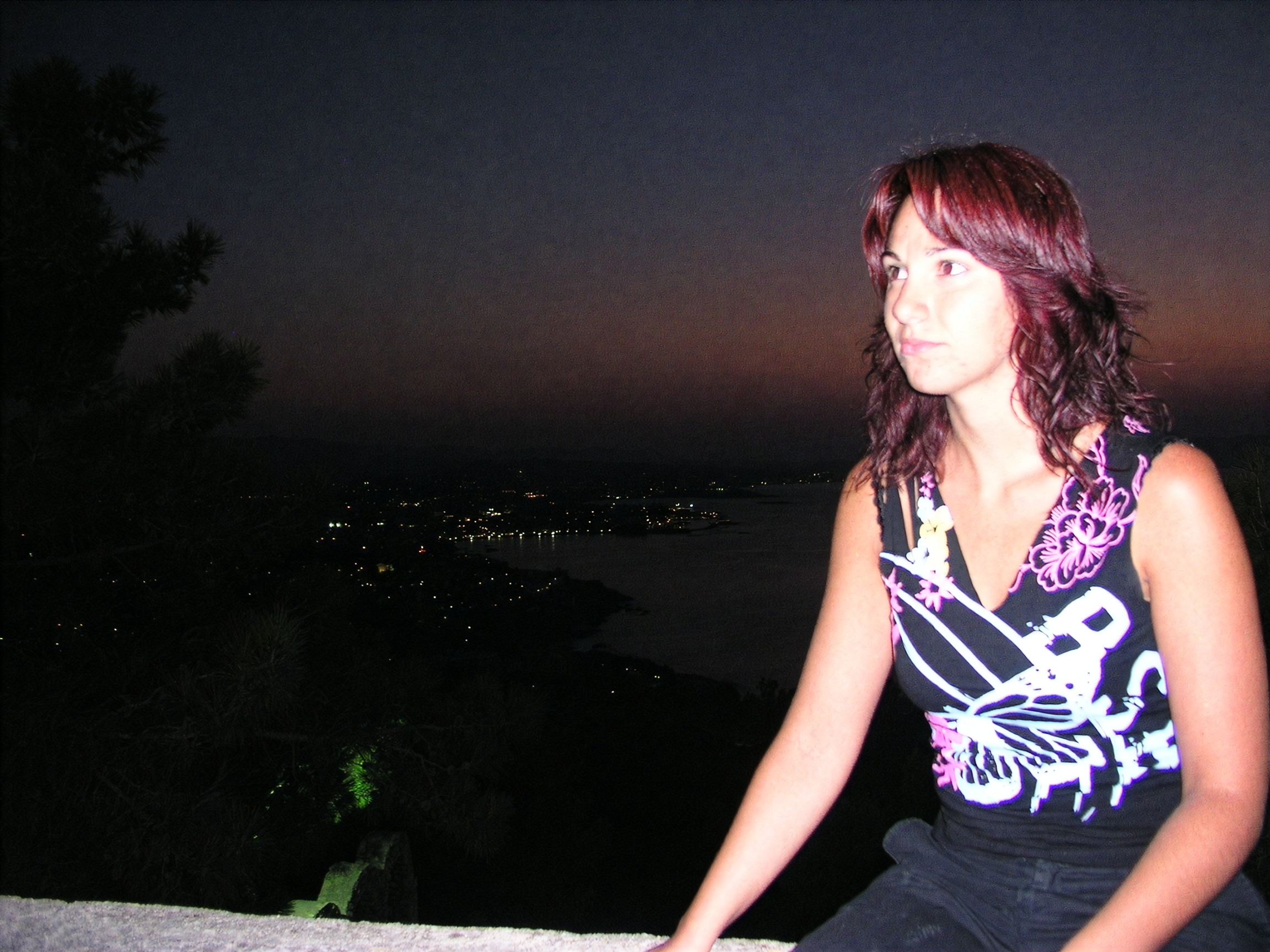} & 
\includegraphics[width=0.18\textwidth]{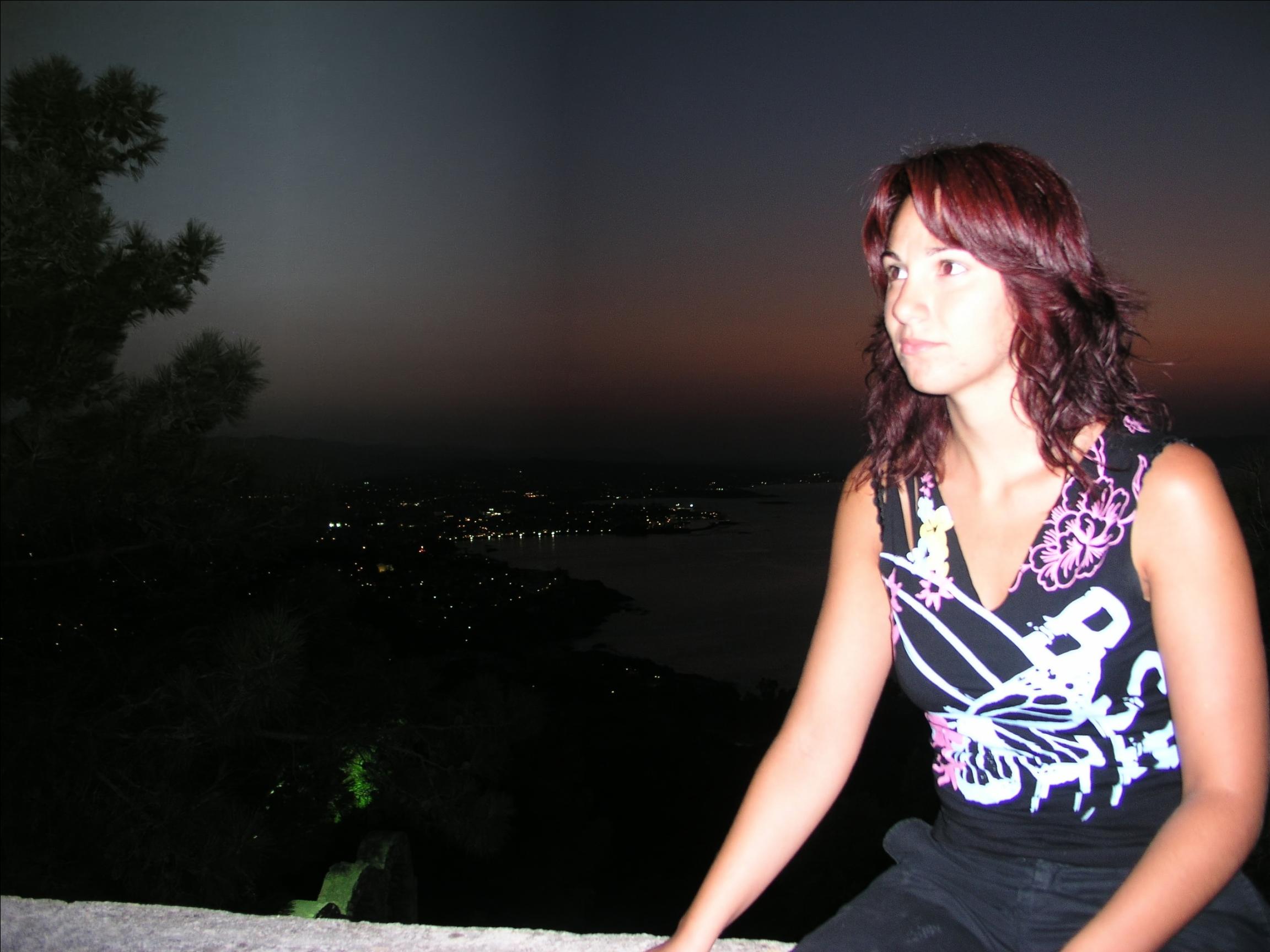} & 
\includegraphics[width=0.18\textwidth]{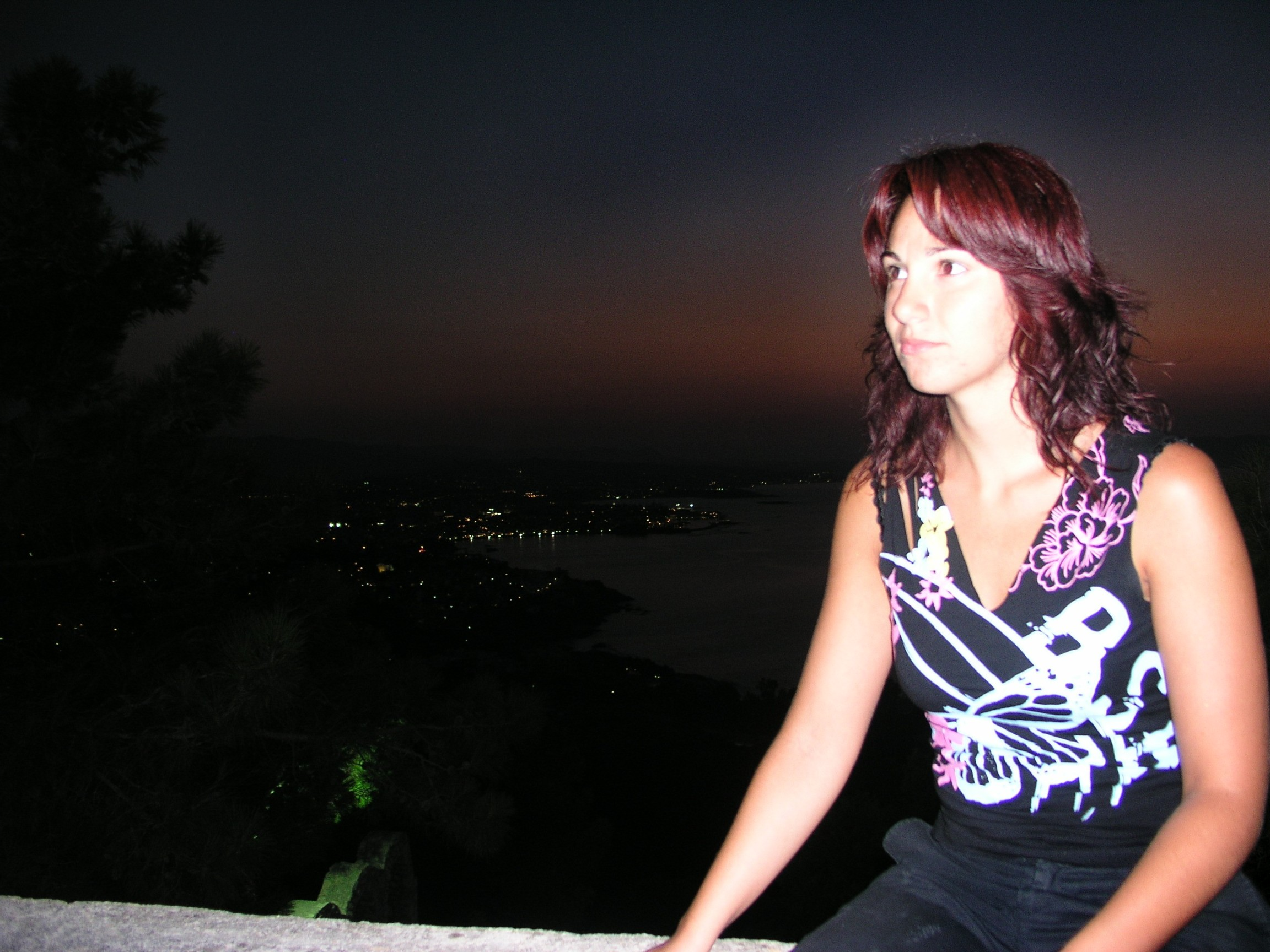} & 
\includegraphics[width=0.18\textwidth]{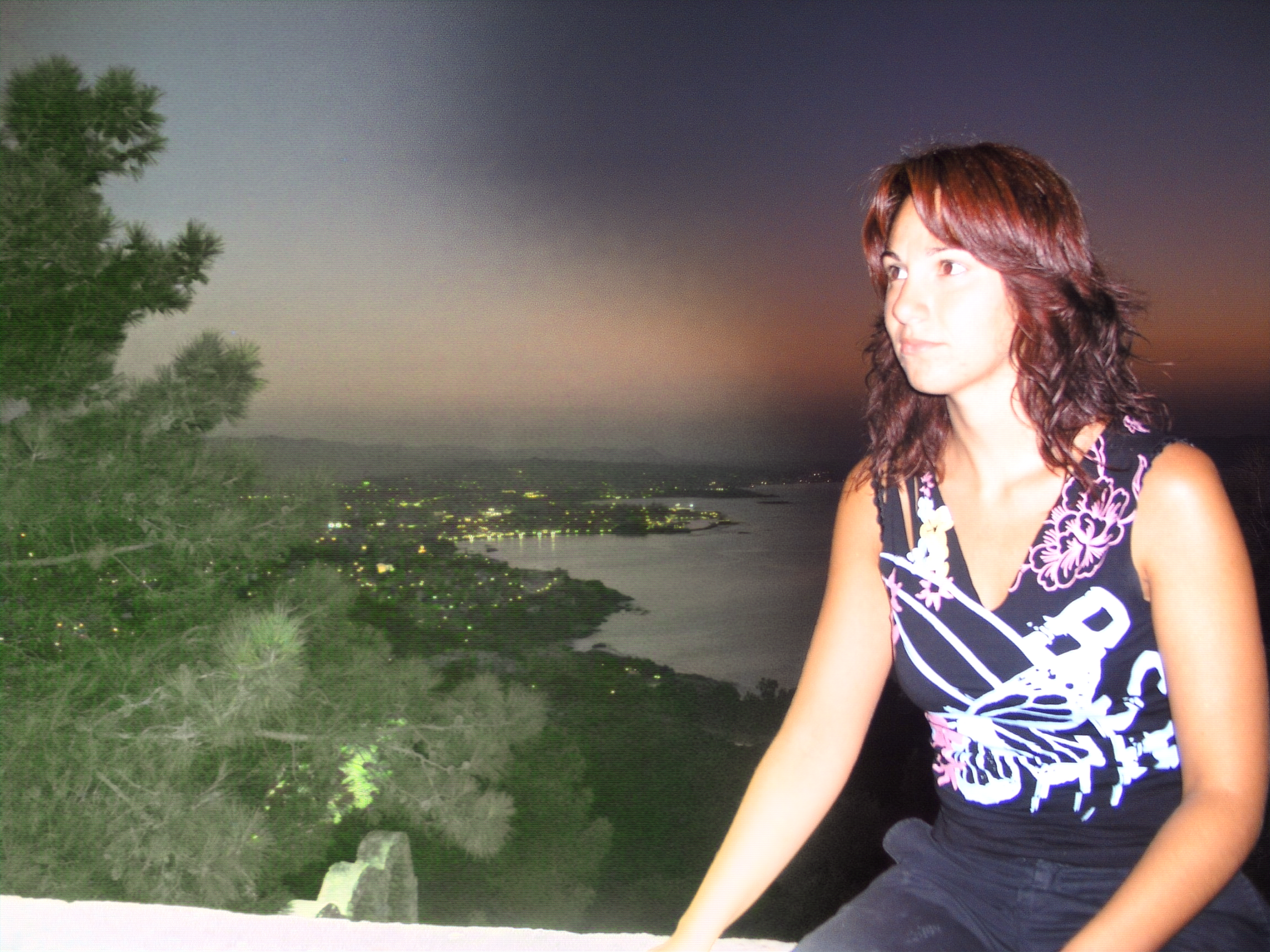} 
\\
Input (LQ) & UniRestore~\cite{chen2025unirestore} & FoundIR~\cite{li2024foundir} & DCPT~\cite{hu2025universal} & \textbf{MaskDCPT (Ours)}
\end{tabular}
\caption{\small \textbf{\textit{Visual comparison on real-world restoration scenarios.}} Zoom in for best view.}
\label{fig:real_world}
\end{figure*}

\noindent \textbf{[ZS] Unseen degradation types: real-world blur and weather}. Table~\ref{tab:real_world} and Figure~\ref{fig:real_world} illustrate the superior generalization capabilities of MaskDCPT in real-world scenarios, significantly outperforming all-in-one restoration methodologies. According to the quantitative metrics, MaskDCPT attained the majority of the state-of-the-art results, notably achieving the lowest Fréchet Inception Distance (FID) in the motion blur task, and delivered superior performance across the other five real-world environments. The visual output delineates that \textbf{(1)} the methods grounded in degradation classification (DCPT~\cite{hu2025universal} and our MaskDCPT) are adept at eliminating small disturbances, \textit{e.g.}, rain and snow from images, unlike the Diffusion-based methods~\cite{chen2025unirestore,li2024foundir}. \textbf{(2)} The integration of mask processing enhances the model's capacity to discern and ameliorate localized degradations. Although DCPT~\cite{hu2025universal} is effective in globally removing rain and snow, it cannot address local low-light conditions. Our MaskDCPT adeptly resolves this problem by accurately illuminating these regions. {More visual comparisons are shown in the supplementary.}


\begin{table*}[ht]
\centering
\fontsize{6.6pt}{\baselineskip}\selectfont
\setlength\tabcolsep{3pt}
\begin{tabular}{c c | c c c c c | c c c}
\toprule[0.15em]
{\textbf{Dataset}} & {\textbf{Method}} & DeblurGAN & DeblurGANv2 & SRN & DMPHN & Restormer & DegAE-Restormer~\cite{Liu_2023_DegAE} & DCPT-Restormer~\cite{hu2025universal} & \textbf{MaskDCPT-Restormer (Ours)} \\
\midrule[0.15em]
\multirow{2}{*}{\textbf{GoPro}} & PSNR $\uparrow$ &  28.70 & 29.55 & 30.26 & 31.20 & 32.92 & 33.03 (\textbf{+0.11}) & 33.12 (\textbf{+0.20}) & 33.29 (\textbf{+0.37}) \\
& SSIM $\uparrow$ &  0.858 & 0.934 & 0.934 & 0.940 & 0.961 & - & 0.962 & 0.964 (\textbf{+0.03}) \\
\midrule[0.1em]
\multirow{2}{*}{\textbf{HIDE}} & PSNR $\uparrow$ & 24.51 & 26.62 & 28.36  & 29.09 & 31.22 & 31.43 (\textbf{+0.21}) & 31.47 (\textbf{+0.25}) & 31.55 (\textbf{+0.33}) \\
 & SSIM $\uparrow$ &  0.871 & 0.875 & 0.915 & 0.924 & 0.942 & - & 0.946 (\textbf{+0.04}) & 0.946 (\textbf{+0.04}) \\
\midrule[0.15em]
\midrule[0.15em]
{\textbf{Dataset}} & {\textbf{Method}} & SIRR & MSPFN & LPNet & AirNet & Restormer & DegAE-Restormer~\cite{Liu_2023_DegAE} & DCPT-Restormer~\cite{hu2025universal} & \textbf{MaskDCPT-Restormer (Ours)} \\ 
\midrule[0.15em]
\multirow{2}*{\textbf{Test100}} & PSNR $\uparrow$ & 32.37 & 33.50 & 33.61 & 34.90 & 36.74 & {35.39} (\textbf{-1.35}) & {{37.24}} (\textbf{+0.50}) & {{37.70}} (\textbf{+0.96}) \\
& SSIM $\uparrow$ & 0.926 & 0.948 & 0.958 & 0.977 & 0.978 & {0.972} (\textbf{-0.06}) & {{0.980}} (\textbf{+0.02}) & {{0.984}} (\textbf{+0.06}) \\ 
\bottomrule[0.15em]
\end{tabular}
\caption{\small \textbf{[FT] \textit{Single Image Motion Deblurring results}} in the single-task setting on the GoPro dataset. \textbf{\textit{Image Deraining results}} in the single-task setting on the Test100 dataset.}\label{table:single}
\vspace{-5mm}
\end{table*}

\noindent \textbf{[FT] Single-task degradation: motion blur and rain}. MaskDCPT is suitable for pre-training on a single task. Table~\ref{table:single} shows that Restormer pre-trained with MaskDCPT outperforms 0.37 dB on GoPro. MaskDCPT remains an appropriate approach for pre-training on image deraining tasks. In contrast, DegAE~\citep{Liu_2023_DegAE} exhibits a reduced performance in image deraining. MaskDCPT exhibits greater universality.

\subsection{Image restoration on mixed degradation}

\noindent \textbf{Dataset}. We use the UIR-2.5M-mixed as a training dataset, and test our model on CDD and LoL-Blur. {The testset comprises prevalent degradation combinations, including low-light, haze, rain-streaks, and snow. We conduct evaluations exclusively on three-mixed degradation to illustrate the advantages of MaskDCPT in restoring intricate degradation mixtures.}

\noindent \textbf{Implementation details}. We use the NAFNet pre-trained by MaskDCPT on 5D all-in-one restoration datasets to initialize model. We use the AdamW optimizer with the initial learning rate $3 \times 10^{-4}$ gradually reduced to $1 \times 10^{-6}$ with the cosine annealing schedule to train our image restoration models. The training runs for 750k iters with batch size 32 on 4 NVIDIA L40 GPUs.

\begin{table*}[ht]
\centering
\fontsize{6.6pt}{\baselineskip}\selectfont
\setlength\tabcolsep{3pt}
\begin{tabular}{l|c|c|c}
\toprule[0.15em]
Method & L + H + RS & L + H + S & L + B + N  \\ 
\midrule[0.15em]
DACLIP~\cite{luo2023controlling} & 25.86 / 0.797 / 0.210 / 25.12 & 25.22 / 0.800 / 0.205 / 28.87 & 26.45 / 0.862 / 0.147 / 11.48 \\
InstructIR~\cite{conde2024instructir} & 24.84 / 0.777 / 0.233 / 28.71 & 24.32 / 0.760 / 0.279 / 40.33 & 26.33 / 0.860 / 0.163 / 17.31 \\
OneRestore~\cite{guo2024onerestore} & 25.18 / 0.799 / 0.165 / 24.85 & 25.28 / 0.802 / 0.148 / 24.90 & 25.02 / 0.788 / 0.227 / 34.32 \\
MoceIR~\cite{zamfir2025complexity} & 25.41 / 0.801 / 0.213 / 30.04 & 25.40 / 0.802 / 0.208 / 28.51 & 24.28 / 0.753 / 0.317 / 43.28 \\
DCPT-NAFNet~\cite{hu2025universal} & 25.76 / \textbf{0.817} / 0.188 / 23.01 & 25.90 / \textbf{0.819} / 0.174 / 23.56 & 25.56 / 0.829 / 0.181 / 23.69 \\ 
\textbf{MaskDCPT-NAFNet (Ours)} & \textbf{26.24} / 0.811 / \textbf{0.151} / \textbf{22.80} & \textbf{26.38} / 0.814 / \textbf{0.143} / \textbf{22.90} & \textbf{27.13} / \textbf{0.881} / \textbf{0.131} / \textbf{10.27} \\ 
\bottomrule[0.15em]
\end{tabular}
\caption{\small \textbf{\textit{Mixed degraded image restoration results}} on CDD~\citep{guo2024onerestore} and LoL-Blur~\cite{zhou2022lednet} in terms of PSNR$\uparrow$ / SSIM$\uparrow$ / LPIPS$\downarrow$ / FID$\downarrow$. All methods are trained on UIR-2.5M-mixed to ensure fair comparison.}\label{tab:mixed}
\vspace{-2mm}
\end{table*}

\begin{figure*}[ht]
\centering
\setlength{\tabcolsep}{2pt}
\begin{tabular}{ccccc}
\includegraphics[width=0.18\textwidth]{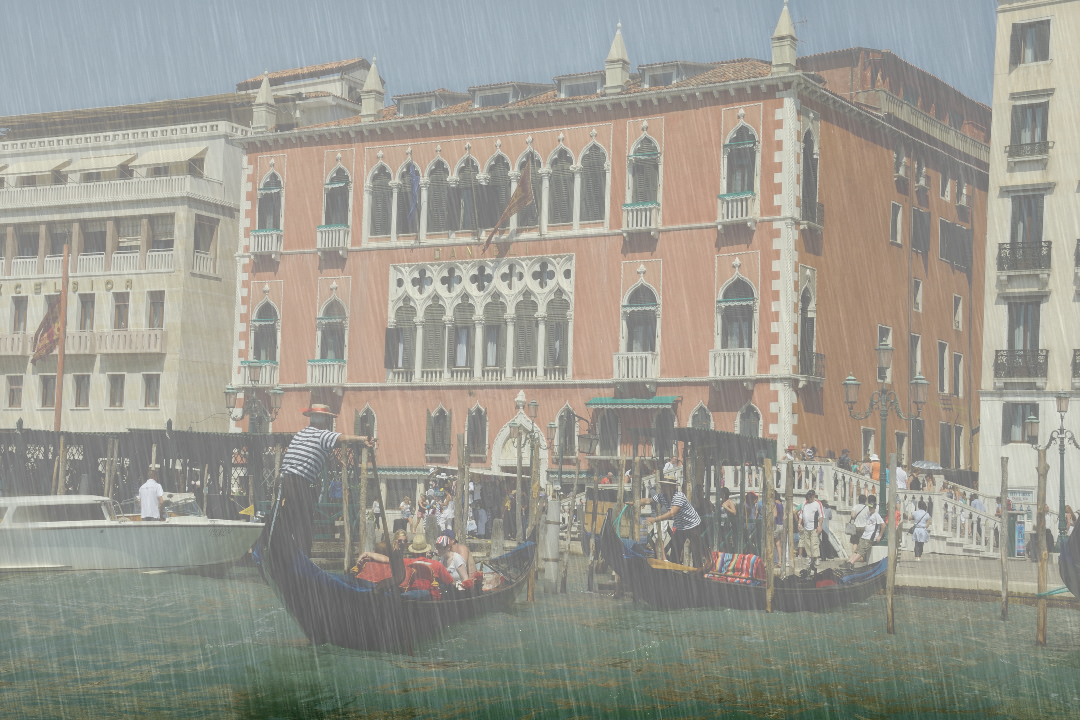} & 
\includegraphics[width=0.18\textwidth]{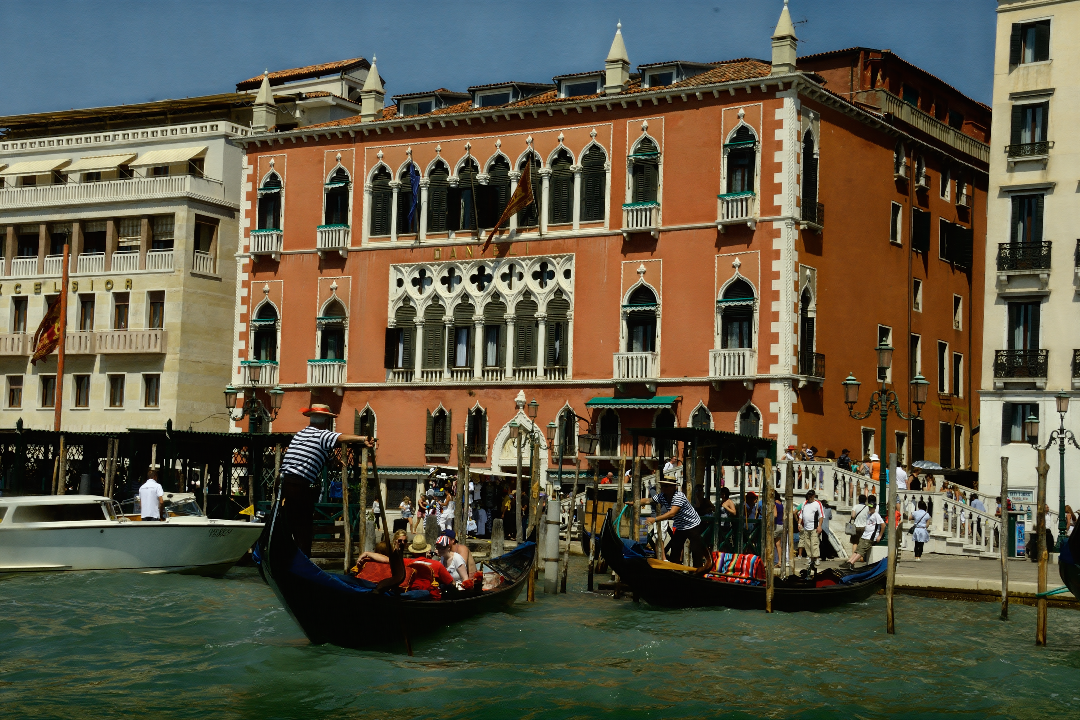} & 
\includegraphics[width=0.18\textwidth]{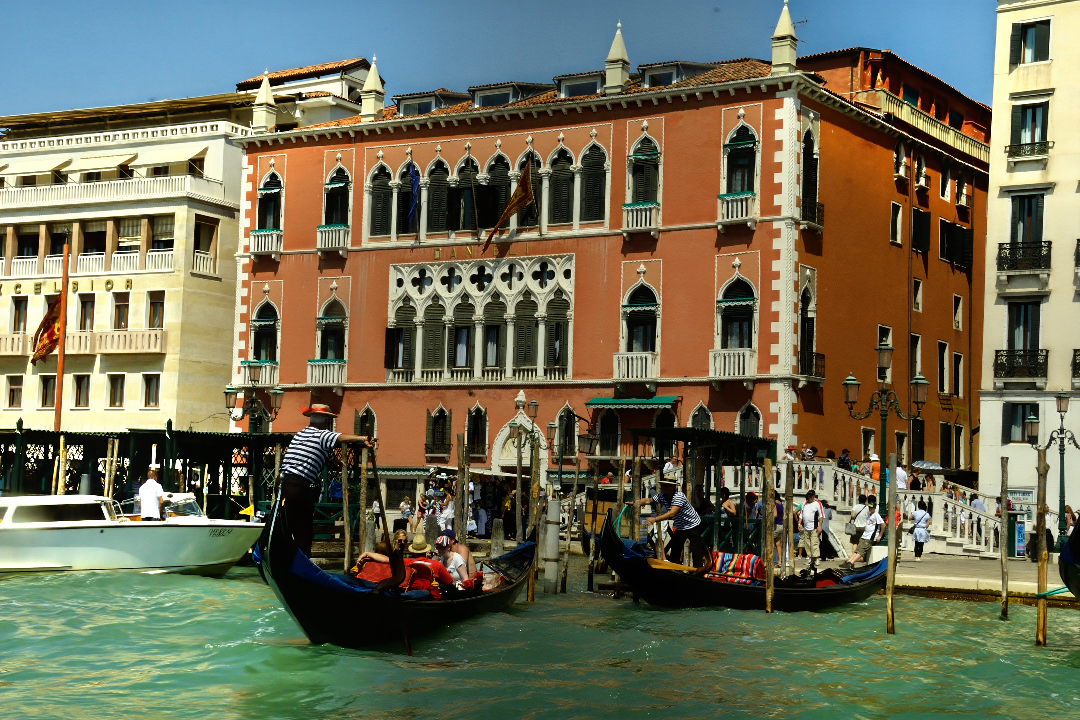} & 
\includegraphics[width=0.18\textwidth]{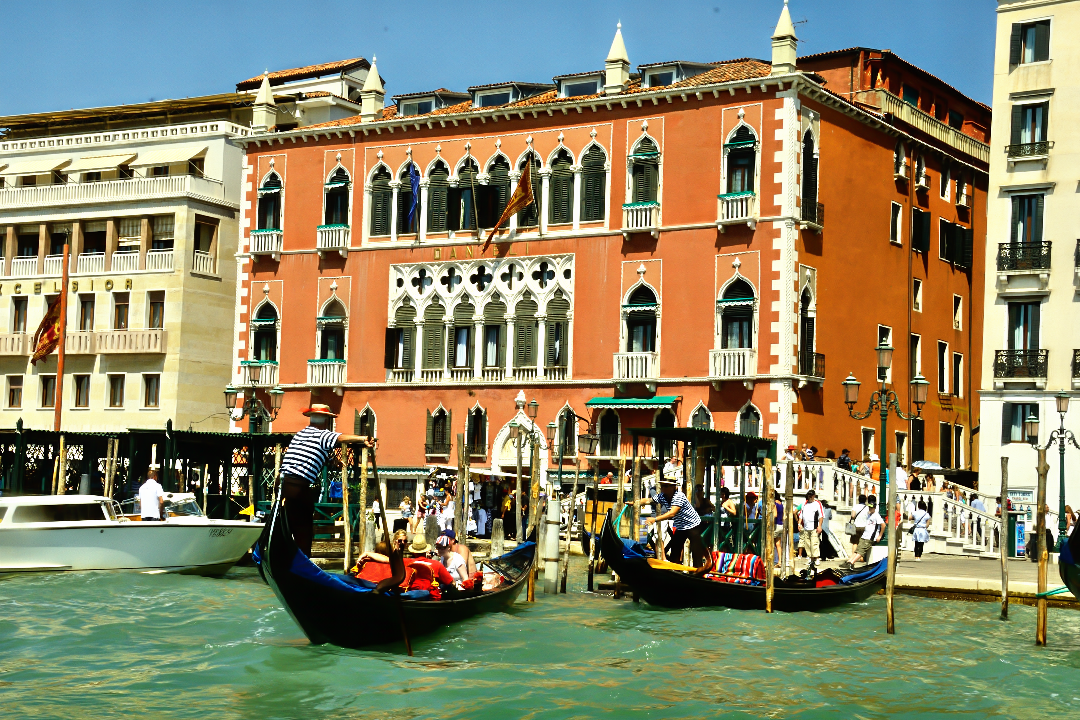} & 
\includegraphics[width=0.18\textwidth]{figures/cdd_compare/maskdcpt.png} 
\\
Input (LQ) & MoceIR~\cite{zamfir2025complexity} & DCPT~\cite{hu2025universal} & \textbf{MaskDCPT (Ours)} & Target (HQ)
\end{tabular}
\caption{\small \textbf{\textit{Visual comparison on mixed degradation scenarios.}} MaskDCPT can restore the illumination globally.}
\label{fig:mixed}
\end{figure*}

\noindent \textbf{Results of restoration on mixed degradation} are displayed in Table~\ref{tab:mixed}. MaskDCPT can deliver substantial performance enhancements to the restoration model in mixed degradation scenarios. Compared with OneRestore~\cite{guo2024onerestore}, MaskDCPT demonstrates a PSNR improvement of 1.06 dB for mixed degradations involving low-light, haze and rain degradation, and 2.11 dB for those involving low-light, blur, and noise degradation. For compelling evidence, Figure~\ref{fig:mixed} provides a visual comparison of image restoration in three composite degradation samples (low-light + haze + rain). NAFNet pre-trained with MaskDCPT can restore more natural result from mixed-degraded image and fully preserve image texture and detail such as lighting and building textures.

\subsection{Ablation studies}

Our conference paper~\cite{hu2025universal} has demonstrated the necessity of decoder architecture, multi-scale feature extraction, training stages, and pre-training it self in DCPT through the performance of several ablation experiments. We hereby comprehensively analyze our newly added mask mechanism. The ablations are performed with PromptIR~\citep{promptir} in the 5D all-in-one image restoration task, in terms of PSNR $\uparrow$.

\begin{table*}[!ht]
\begin{minipage}[t]{0.48\textwidth}
\centering
\fontsize{6.6pt}{\baselineskip}\selectfont
\setlength\tabcolsep{3pt}
\setlength{\tabcolsep}{3pt}
\begin{tabular}{c|c|c|c|c|c}
\toprule[0.15em]
Mask ratio (\%) & Dehazing & Deraining & Denoising & Debluring & Low-light \\
\midrule[0.15em]
0 & 30.93 & 37.18 & 31.27 & 28.86 & 23.09 \\
\midrule
25 & 31.93 & 38.84 & 31.28 & 29.10 & 25.71 \\
\midrule
\textbf{50} & \textbf{32.71} & \textbf{39.12} & \textbf{31.30} & \textbf{29.99} & \textbf{26.30} \\
\midrule
75 & 32.66 & 39.08 & 31.24 & 29.84 & 26.00 \\
\bottomrule[0.15em]
\end{tabular}
\caption{Ablations of the mask ratio.}\label{tab:mask_ratio}
\vspace{-3mm}
\end{minipage}
\begin{minipage}[H]{0.01\textwidth}
\end{minipage}
\begin{minipage}[t]{0.48\textwidth}
\centering
\fontsize{6.6pt}{\baselineskip}\selectfont
\setlength\tabcolsep{3pt}
\setlength{\tabcolsep}{3pt}
\begin{tabular}{c|c|c|c|c|c}
\toprule[0.15em]
Patch size & Dehazing & Deraining & Denoising & Debluring & Low-light \\
\midrule[0.15em]
1 & 29.33 & 36.12 & 31.09 & 27.89 & 23.37 \\
\midrule
4 & 31.99 & 39.03 & 31.28 & 29.74 & 26.11 \\
\midrule
\textbf{16} & \textbf{32.71} & \textbf{39.12} & \textbf{31.30} & \textbf{29.99} & \textbf{26.30} \\
\midrule
32 & 32.17 & 38.88 & 31.29 & 29.68 & 25.75 \\
\bottomrule[0.15em]
\end{tabular}
\caption{Ablations of masked patch size.}\label{tab:patch_size}
\vspace{-3mm}
\end{minipage}
\end{table*}

\noindent \textbf{Impact of mask ratio.} As shown in Table~\ref{tab:mask_ratio}, it was observed that selecting a mask ratio of 50\% optimizes restoration performance. In contrast, when the mask ratio is reduced to 0, MaskDCPT reverts to DCPT~\cite{hu2025universal}, thus losing its ability to train simultaneously for degradation discrimination and image reconstruction, resulting in a notable performance decline.

\noindent \textbf{Impact of masked patch size.} Refer to Table~\ref{tab:patch_size}, a patch size of 16 is optimal for MaskDCPT. In instances where the patch size is adjusted to 1, the masking approach aligns with RAM~\cite{qin2024restore}. Our findings indicate that employing a patch size of 1 during pre-training with degradation classification can notably diminish restoration performance. This occurs because pixel-level masks disrupt the distribution of degradation information throughout the image, thereby impeding the model's ability to effectively detect degradation, which ultimately impacts the restoration performance.

\begin{table}[ht]
\centering
\fontsize{6.6pt}{\baselineskip}\selectfont
\setlength\tabcolsep{3pt}
\setlength{\tabcolsep}{3pt}
\begin{tabular}{c|c|c|c|c|c}
\toprule[0.15em]
Masking method & Dehazing & Deraining & Denoising & Debluring & Low-light \\
\midrule[0.15em]
square & 32.18 & 38.94 & 30.84 & 29.07 & 25.82 \\
\midrule
block-wise & 32.22 & 39.00 & 30.98 & 29.27 & 25.90 \\
\midrule
\textbf{random} & \textbf{32.71} & \textbf{39.12} & \textbf{31.30} & \textbf{29.99} & \textbf{26.30} \\
\bottomrule[0.15em]
\end{tabular}
\caption{Ablations of masking methods.}\label{tab:mask_method}
\vspace{-3mm}
\end{table}

\noindent {\textbf{Impact of masking methods.} Following SimMIM~\cite{xie2022simmim}, we conducted ablations involving a variety of masking methods. As evidenced by the results presented in Table~\ref{tab:mask_method}, the random masking strategy exerts optimal performance in the 5D all-in-one image restoration task. This observation can be attributed to the inherently pixel-intensive nature of image restoration tasks, which necessitate the model's proficiency in processing various image regions. The application of random masks improves the model’s ability to fit the distribution of pixels between disparate image regions, thus markedly increasing restoration performance.}

\subsection{Discussions}


\begin{table}[ht]
\centering
\fontsize{6.6pt}{\baselineskip}\selectfont
\setlength\tabcolsep{3pt}
\begin{tabular}{c|c|c|c|c|c|c}
\toprule[0.15em]
Method & MaskDCPT iterations & 0 & 25k & 50k & 75k & 100k \\
\midrule[0.15em]
\multirow{2}{*}{SwinIR~\cite{liang2021swinir}} & Initial DC Acc. (\%) & 54 & 69 & 82 & 89 & 94 \\
\specialrule{0pt}{1pt}{1pt}
\cline{2-7}
\specialrule{0pt}{1pt}{1pt}
& PSNR (dB) & 25.04 & 26.10 & 27.58 & 28.44 & 29.21 \\
\midrule
\multirow{2}{*}{NAFNet~\cite{nafnet}} & Initial DC Acc. (\%) & 52 & 90 & 95 & 97 & 98 \\
\specialrule{0pt}{1pt}{1pt}
\cline{2-7}
\specialrule{0pt}{1pt}{1pt}
& PSNR (dB) & 27.77 & 29.40 & 30.95 & 31.88 & 32.09 \\
\midrule
\multirow{2}{*}{Restormer~\cite{Zamir2021Restormer}} & Initial DC Acc. (\%) & 60 & 92 & 97 & 99 & 99 \\
\specialrule{0pt}{1pt}{1pt}
\cline{2-7}
\specialrule{0pt}{1pt}{1pt}
& PSNR (dB) & 27.60 & 29.34 & 30.76 & 31.63 & 31.98 \\
\midrule
\multirow{2}{*}{PromptIR~\cite{promptir}} & Initial DC Acc. (\%) & 56 & 90 & 94 & 97 & 98 \\
\specialrule{0pt}{1pt}{1pt}
\cline{2-7}
\specialrule{0pt}{1pt}{1pt}
& PSNR (dB) & 28.11 & 29.77 & 30.43 & 31.50 & 31.88 \\
\bottomrule[0.15em]
\end{tabular}
\caption{All-in-one restoration performance improved as the degradation classification accuracy increased. The PSNR are averaged among 5 tasks in 5D all-in-one restoration. ``DC" donates the degradation classification.}\label{tab:acc_restoration}
\vspace{-3mm}
\end{table}

\noindent {\textbf{Restoration performance as the degradation classification accuracy changes.} The results demonstrate a direct correlation between enhancements in the accuracy of degradation classification during pre-training and subsequent improvements in the network's all-in-one restoration performance. As delineated in Table~\ref{tab:acc_restoration}, there is a notable improvement in the performance of restoration models, concomitant with an increase in the initial degradation classification accuracy. This correlation implies that the effectiveness of MaskDCPT is largely attributable to its facilitation of degradation classification prior to the initiation of restoration training.}

\begin{table}[!ht]
\centering
\fontsize{6.6pt}{\baselineskip}\selectfont
\setlength\tabcolsep{3pt}
\begin{tabular}{c|l|c|c}
\toprule[0.15em]
Types & Methods & w/o MaskDCPT & w MaskDCPT \\
\midrule[0.15em]
\multirow{4}*{5D} & Restormer~\cite{Zamir2021Restormer} & 27.60 / 0.112 & 31.98 / 0.055 \\
~ & \quad + instructs~\cite{conde2024instructir} & 30.11 / 0.083 & 31.93 / 0.056 \\
~ & \quad + frequency~\cite{cui2024adair} & 30.09 / 0.089 & 31.99 / 0.055 \\
~ & \quad + MoE~\cite{zamfir2025complexity} & 30.62 / 0.079 & 32.07 / 0.050 \\
\midrule[0.15em]
\multirow{4}*{9D} & Restormer~\cite{Zamir2021Restormer} & 27.14 / 0.139 & 31.37 / 0.058 \\
~ & \quad + instructs~\cite{conde2024instructir} & 29.67 / 0.094 & 31.40 / 0.059 \\
~ & \quad + frequency~\cite{cui2024adair} & 29.20 / 0.118 & 31.32 / 0.061 \\
~ & \quad + MoE~\cite{zamfir2025complexity} & 29.46 / 0.101 & 31.43 / 0.056 \\
\midrule[0.15em]
\multirow{4}*{12D} & Restormer~\cite{Zamir2021Restormer} & 26.88 / 0.196 & 30.79 / 0.061 \\
~ & \quad + instructs~\cite{conde2024instructir} & 29.03 / 0.140 & 30.80 / 0.061 \\
~ & \quad + frequency~\cite{cui2024adair} & 28.26 / 0.173 & 30.74 / 0.063 \\
~ & \quad + MoE~\cite{zamfir2025complexity} & 28.71 / 0.159 & 30.91 / 0.058 \\
\bottomrule[0.15em]
\end{tabular}
\caption{5D (H, RS, GN, MB, LL) restoration performance in terms of PSNR $\uparrow$ / LPIPS $\downarrow$ of degradation-aware architectures as influenced by training methods and scaled degradation types.}
\vspace{-3mm}
\label{tab:discussion}
\end{table}

\noindent \textbf{What do degradation-aware architectures bring to?} MaskDCPT has been shown to be highly effective in improving model performance. Furthermore, a baseline model trained with MaskDCPT demonstrates superior performance compared to degradation-aware architectures. We wonder: ``What do degradation-aware architectures bring to restoration performance?" We examined variations in the restoration performance of three degradation-aware architectures: instruction~\cite{conde2024instructir}, frequency~\cite{cui2024adair}, and Mixture of Experts (MoE)~\cite{zamfir2025complexity} as influenced by changes in training methods and increased degradation types. Restormer~\cite{Zamir2021Restormer} serves as the baseline model. The experimental results are presented in Table~\ref{tab:discussion}. \textbf{(1)} When training models from scratch, degradation-aware architectures can provide specific performance enhancements. \textbf{(2)} However, as the types of degradation increase, the performance of the model experiences varying degrees of decline. Both MaskDCPT and degradation-aware architectures can mitigate this performance decline. From a network architecture perspective, the instruction-based methodology emerges as the most effective means of alleviating this performance decline. \textbf{(3)} After training with MaskDCPT, degradation-aware architectures consistently approximate the baseline performance. It suggests that such designs may not be able to increase the model performance ceiling after fully converged training.



\section{Conclusion}

This paper first validates that randomly initialized restoration models achieve baseline degradation classification performance while preserving robustness with masked input images. Furthermore, models trained for all-in-one restoration exhibit superior classification accuracy. To enhance this efficacy and robustness, we introduce MaskDCPT and experimentally demonstrate its effectiveness in universal image restoration. By integrating degradation classification priors with image distribution learning, MaskDCPT enhances a 3-4 dB PSNR gain in all-in-one restoration and 35 \% less PIQE in real-world scenarios for restoration models. In addition, we gather an extensive dataset for universal image restoration, UIR-2.5M, which is able to improve the generalization of restoration models when addressing unseen degradation. In future work, efforts will be directed to improve the generalization of restoration models in the presence of unseen and complex degradation.




\bibliography{reference}
\bibliographystyle{unsrt}


 




\vfill

\end{document}